\documentclass{article}

\usepackage{hyperref}
\usepackage{url}

\usepackage{times}
\usepackage{latexsym}

\usepackage[T1]{fontenc}
\usepackage[utf8]{inputenc}

\usepackage{microtype}

\usepackage{inconsolata}

\usepackage{graphicx}
\usepackage{float}
\usepackage{amsmath,gensymb,amssymb}
\usepackage{bbm}
\usepackage{subcaption}

\usepackage{multirow}
\usepackage{tabularx}
\usepackage{setspace}

\usepackage{wrapfig}
\usepackage[table]{xcolor}
\usepackage{xspace}

\usepackage{tcolorbox}
\usepackage[utf8]{inputenc} 
\usepackage[shortlabels]{enumitem}
\setlist[itemize]{leftmargin=*}
\setlist[enumerate]{leftmargin=*}
\usepackage[normalem]{ulem}

\usepackage{booktabs}
\usepackage{soul}
\usepackage{todonotes}

\newcommand{\boldstart}[1]{\noindent\textbf{#1}}

\newcommand{\env}{\texttt{$\langle$ENV$\rangle$}\xspace}
\newcommand{\lan}{\texttt{$\langle$LAN$\rangle$}\xspace}

\definecolor{tticblue}{RGB}{0, 94, 184}
\usepackage{array}
\newcolumntype{C}[1]{>{\centering\arraybackslash}m{#1}}
\newcolumntype{R}{>{\raggedleft\arraybackslash}p{1cm}}

\usepackage{booktabs}
\usepackage{rotating}

\usepackage{epigraph}
\newrobustcmd*{\citefirstlastauthor}{\AtNextCite{\DeclareNameAlias{labelname}{given-family}}\citeauthor}
\AtBeginDocument{}

\usepackage{tabularx}
\usepackage{graphicx}
\usepackage{booktabs}
\usepackage{array}

\usepackage[accepted]{icml2026}

\icmltitlerunning{Emergence of Symbol Grounding}

\begin{document}

\twocolumn[

\icmltitle{The Mechanistic Emergence of Symbol Grounding in Language Models}

% unless you've provided the [accepted] option to the icml2026


\icmlsetsymbol{colead}{*}
\icmlsetsymbol{coadvise}{$\dagger$}

\begin{icmlauthorlist}

\icmlauthor{Shuyu Wu}{colead,umich}
\icmlauthor{Ziqiao Ma}{colead,umich}
\icmlauthor{Xiaoxi Luo}{colead,waterloo,vector}
\icmlauthor{Yidong Huang}{umich,unc}
\icmlauthor{Josue Torres-Fonseca}{umich}

\icmlauthor{Freda Shi}{coadvise,waterloo,vector}
\icmlauthor{Joyce Chai}{coadvise,umich}

\end{icmlauthorlist}

\icmlaffiliation{umich}{University of Michigan}
\icmlaffiliation{unc}{University of North Carolina at Chapel Hill}
\icmlaffiliation{waterloo}{University of Waterloo}
\icmlaffiliation{vector}{Vector Institute}

\icmlcorrespondingauthor{Shuyu Wu}{shuyuwu@umich.edu}
\icmlcorrespondingauthor{Freda Shi}{fhs@uwaterloo.ca}
\icmlcorrespondingauthor{Joyce Chai}{chaijy@umich.edu}

\icmlkeywords{Machine Learning, ICML}




\vskip 0.3in
]

\printAffiliationsAndNotice{\icmlEqualContribution} 

\begin{abstract}

%Do language models acquire symbol grounding in \citet{harnad1990symbol}’s sense, that is, non-arbitrary, causally useful links between symbols and referents? 
Symbol grounding~\citep{harnad1990symbol} describes how symbols such as words acquire their meanings by connecting to real-world sensorimotor experiences.
Recent work has shown preliminary evidence that grounding may emerge in (vision-)language models trained at scale without using explicit grounding objectives. 
% However, where in these models and what potential mechanisms give rise to the emergence are not explored. 
Yet, the specific loci of this emergence and the mechanisms that drive it remain largely unexplored.
To address this problem, we introduce a controlled evaluation framework that systematically traces how symbol grounding arises within the internal computations through mechanistic and causal analysis.
Our findings show that grounding concentrates in middle-layer computations and is implemented through the aggregate mechanism, where attention heads aggregate the environmental ground to support the prediction of linguistic forms. 
This phenomenon replicates in multimodal dialogue and across architectures (Transformers and state-space models), but not in unidirectional LSTMs. 
Our results provide behavioral and mechanistic evidence that symbol grounding can emerge in language models, with practical implications for predicting and potentially controlling the reliability of generation.

%assigns each concept two distinct tokens: one appearing in non-verbal scene descriptions and another in linguistic utterances. 
%This separation prevents trivial identity mappings and enables direct tests of grounding. 
%Behaviorally, models trained from scratch show consistent surprisal reduction when the linguistic form is preceded by its matching scene token, relative to matched controls, and this effect cannot be fully explained by co-occurrence statistics.
%Mechanistically, saliency flow and tuned-lens analyses converge on the finding that grounding concentrates in middle-layer computations and is implemented through the aggregate mechanism, where attention heads aggregate the environmental ground to support the prediction of linguistic forms. 
%The phenomenon replicates in multimodal dialogue and across architectures (Transformers and state-space models), but not in unidirectional LSTMs. 
%Together, these results provide behavioral and mechanistic evidence that symbol grounding can emerge in autoregressive LMs, while delineating the architectural conditions under which it arises.

\end{abstract}
\section{Introduction}
\label{sec:intro}

Symbol grounding \citep{harnad1990symbol} refers to the problem of how abstract and discrete symbols, such as words, acquire meaning by connecting to perceptual or sensorimotor experiences.
In the context of multimodal machine learning, grounding has served as a pre-training objective for vision-language models (VLMs), by connecting linguistic units to the world that gives language meanings \citep{li2022grounded,ma2023world}.
Through supervised fine-tuning with explicit grounding supervision such as entity-phrase mappings, modern VLMs have achieved fine-grained understanding at both region \citep{you2024ferret,peng2024grounding,wang2024cogvlm} and pixel \citep{zhang2024groundhog,rasheed2024glamm,zhang2024omg} levels. 

Meanwhile, with the rising of autoregressive language models \citep[LMs;][\textit{inter alia}]{gpt4o,anthropic2024claude,comanici2025gemini} and their VLM extensions, there is growing interest in identifying and interpreting their emergent capabilities.
Recent work has shown preliminary correlational evidence that grounding may emerge in LLMs \citep{sabet2020simalign,shi-etal-2021-bilingual,wu2025semantic} and VLMs \citep{cao2024emerging,bousselham2024grounding,schnaus2025s} trained at scale, even when solely optimized with the simple next-token prediction objective. 
However, the underlying mechanisms that lead to such an emergence are not well understood. 
To address this limitation, our work seeks to understand the emergence of symbol grounding in multimodal language models, causally and mechanistically tracing how symbol grounding arises within the internal computations.

Given the intractability of directly interpreting large, complex VLMs, we begin by constructing a minimal testbed to investigate grounding mechanisms in a synthetic multimodal setting. 
Our testbed is derived from the CHILDES corpora \citep{macwhinney2000childes}, which provide cognitively plausible contexts and utterances for human language acquisition research. 
In our framework, each word is represented in two distinct forms: one token that appears in non-verbal scene descriptions (e.g., \textit{box} in the environment description) and another that appears in spoken utterances (e.g., \textit{box} in dialogue). 
We refer to these as environmental tokens (\env) and linguistic tokens (\lan), respectively.
A deliberately simple word-level tokenizer assigns separate vocabulary indices to each form, ensuring that they are treated as entirely different tokens by the language model. 
This structural separation prevents grounding from being reduced to the trivial case of shared token identity, and offers a controlled, noise-free proxy for studying the grounding mechanisms.
We use this testbed to derive hypotheses, which are subsequently validated with realistic VLMs.

% To quantify grounding, we can evaluate whether a model trained from scratch can predict the linguistic form based on its environmental counterpart.
% The testbed enables us to draw hypotheses that may subsequently generalize to realistic VLM settings. 

\begin{figure*}[!t]
    \centering
    \begin{subfigure}[t]{.52\textwidth}
        \centering
        \includegraphics[width=1.02\columnwidth]{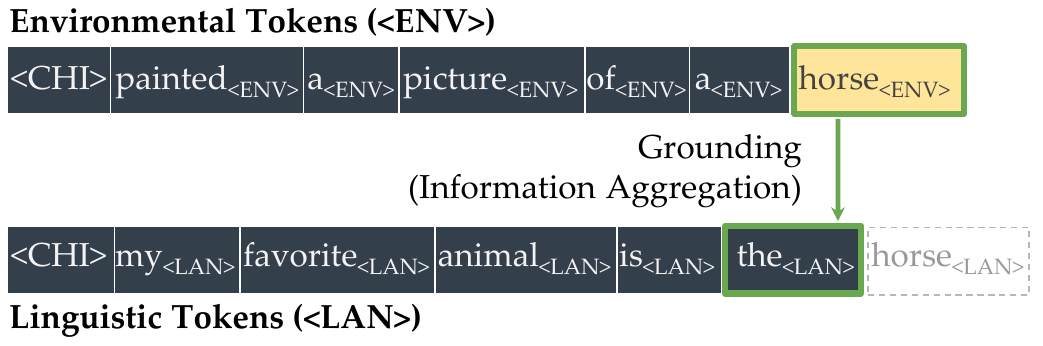}
        \caption{Attention head 8 of layer 7 in GPT-CHILDES.}
        \vspace*{2pt}
        \label{fig:overview_wo_v}
    \end{subfigure}
    ~
    \begin{subfigure}[t]{.46\textwidth}
        \centering
        \includegraphics[width=1.02\columnwidth]{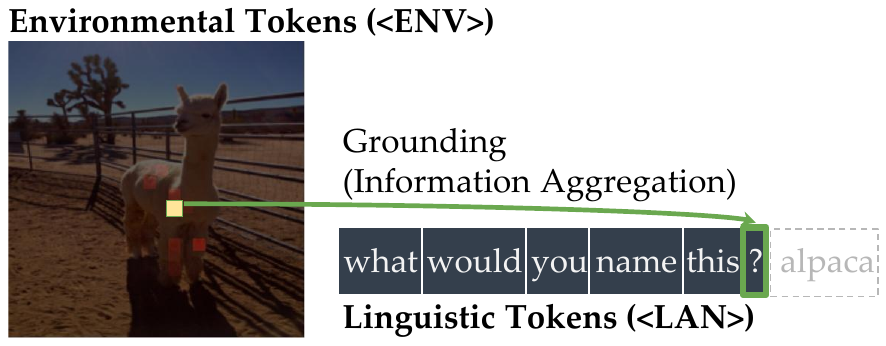}
        \caption{Attention head 7 of layer 20 in LLaVA-1.5-7B.}
        \vspace*{2pt}
        \label{fig:overview_w_v}
    \end{subfigure}
    ~
    \begin{subfigure}[t]{1.0\textwidth}
        \centering
        \includegraphics[width=0.99\columnwidth]{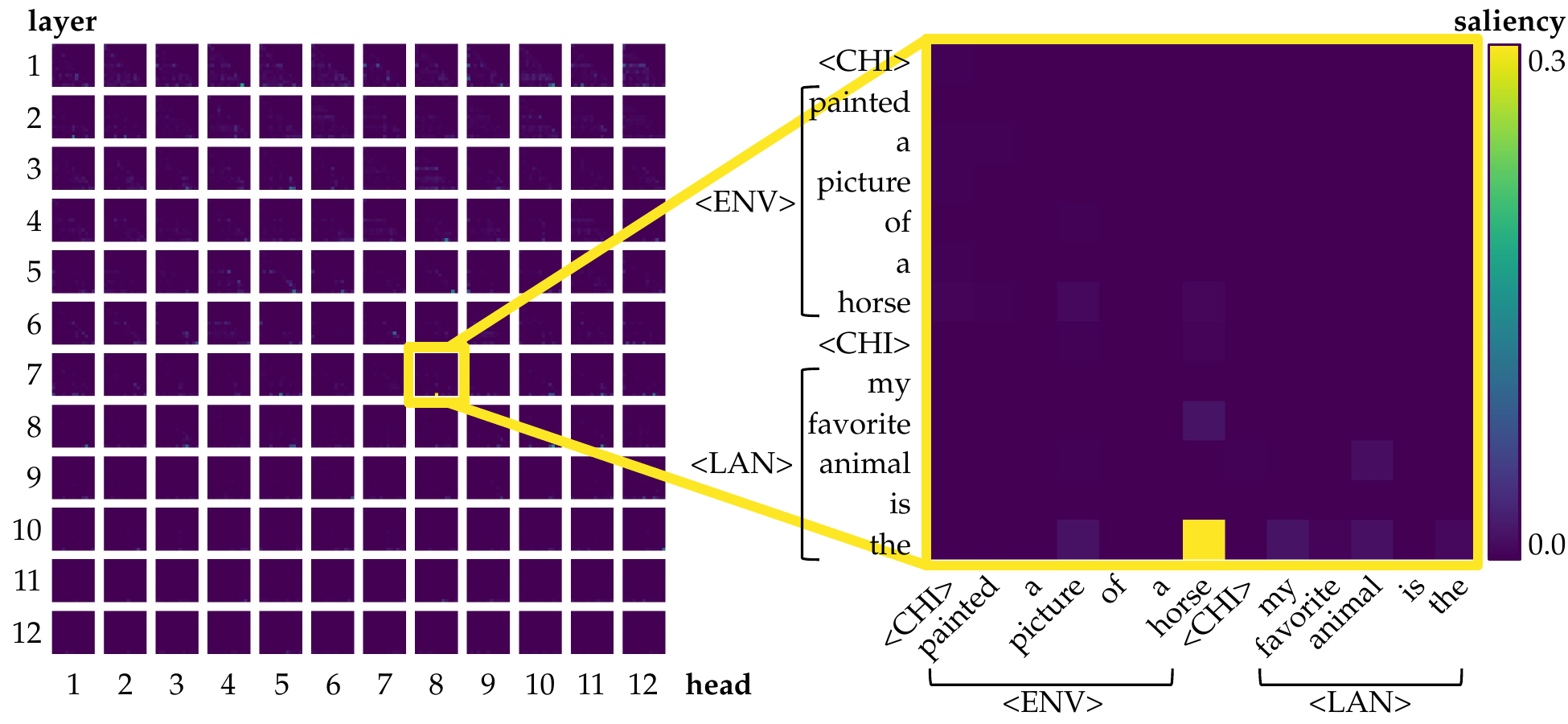}
        \vspace*{2pt}
        \caption{
            Left: saliency over tokens of each head in each layer for the prompt \textit{$\langle$CHI$\rangle$ $\textit{painted}_\env$ $\textit{a}_\env$ $\textit{picture}_\env$ $\textit{of}_\env$ $\textit{a}_\env$ $\textit{horse}_\env$ $\langle$CHI$\rangle$ $\textit{my}_\lan$ $\textit{favorite}_\lan$ $\textit{animal}_\lan$ $\textit{is}_\lan$ $\textit{the}_\lan$}. 
            Right: among all, only one of them (head 8 of layer 7) is identified as an aggregate head, where information flows from $\textit{horse}_\env$ to the current position, encouraging the model to predict $\textit{horse}_\lan$ as the next token.
        }
        \label{fig:overview_demo}
    \end{subfigure}
    \caption{
        Illustration of the symbol grounding mechanism through information aggregation. 
        Lighter colors denote more salient attention, quantified by saliency scores, i.e., gradient $\times$ attention contributions to the loss \citep{wang2023label}. 
        When predicting the next token, aggregate heads \citep{bick2025understanding} emerge to exclusively link environmental tokens (visual or situational context; \env) to linguistic tokens (words in text; \lan). 
        These heads provide a mechanistic pathway for symbol grounding by mapping external environmental evidence into its linguistic form. 
    }
    \label{fig:overview}
\end{figure*}

With the testbed, we quantify grounding using surprisal: specifically, we compare how easily the model predicts a linguistic token (\lan) when its matching environmental token (\env) is present versus when unrelated cues are given instead.
A lower surprisal in the former condition indicates that the model has learned to align environmental grounds with linguistic forms. 
We find that LMs do learn to ground: the presence of environmental tokens consistently reduces surprisal for their linguistic counterparts, in a way that simple co-occurrence statistics cannot fully explain.
To study the underlying mechanisms, we apply saliency analysis \citep{wang2023label} and the tuned lens \citep{belrose2023eliciting}, which converge on the result that grounding relations are concentrated in the middle layers of the network. 
Further analysis of attention heads reveals patterns consistent with the aggregate mechanism~\citep{bick2025understanding}, where attention heads support the prediction of linguistic forms by retrieving their environmental grounds in the context. Figure~\ref{fig:overview_wo_v} and~\ref{fig:overview_demo} illustrates this pattern with an example. 
% Further analysis of attention heads reveals patterns consistent with the gather-and-aggregate mechanism~\citep[G\&A;][]{bick2025understanding}: early heads gather information from environmental tokens, and mid-layer heads aggregate it to support linguistic prediction.

\iffalse 
Our evaluation quantifies grounding behaviorally using surprisal. 
We find that LMs do learn to ground: the presence of environmental tokens consistently lowers surprisal for their linguistic counterparts in ways that cannot be explained by co-occurrence alone.
To examine the mechanisms underlying this behavior, we apply saliency flow analysis and the tuned lens~\citep{belrose2023eliciting}, both of which converge on the finding that grounding relations are concentrated in the middle layers of the network. 
We further analyze attention heads and find patterns consistent with the gather-and-aggregate (G\&A) mechanism~\citep{bick2025understanding}: early heads gather information from environmental tokens, and mid-layer heads aggregate it to support linguistic prediction.
\fi 
% These analyses yield three main findings:
% (1) LMs learn to ground, as the presence of environmental tokens reliably lowers surprisal for their linguistic counterparts beyond what co-occurrence alone can explain;
% (2) grounding effects are localized to mid-layer computations, as revealed by saliency and probing; and
% (3) attention-head behaviors align with the G\&A mechanism previously identified in in-context learning.

Finally, we demonstrate that these findings generalize beyond the minimal testbed with CHILDES data and Transformer models. 
We observe that similar trends appear in a multimodal setting with the Visual Dialog dataset~\citep{das2017visual}, and in state-space models (SSMs) such as Mamba-2~\citep{dao2024transformers}. %, and in hybrid models that combine attention and state-space layers. 
In contrast, we do not observe grounding in unidirectional LSTMs, consistently with their sequential state compression and lack of content-addressable retrieval. 
Taken together, our results show that symbol grounding can mechanistically emerge in autoregressive multimodal LMs, as well as delineating the architectural conditions under which it can arise.

% \jycc{(1) what is symbol grounding,  human like symbol ground; (2) to study grounded language learning, an increasing amount of work..(the previous emergent grounding work).. However, the underlying mechanisms for symbol grounding in LLMs are not understood. (3) to address this issue, this paper.....  (4) brief summary of the findings. }

\vspace*{-2pt}
\section{Related Work}
\label{sec:related}
\vspace*{-2pt}

\boldstart{Language grounding.}
Referential grounding has long been framed as the lexicon acquisition problem: how words map to referents in the world~\citep{harnad1990symbol,gleitman1994acquisition,clark1995lexicon}.
Early work focused on word-to-symbol mappings, designing learning mechanisms that simulate children’s lexical acquisition and explain psycholinguistic phenomena~\citep{siskind1996computational,regier2005emergence,goodman2007bayesian,fazly2010probabilistic}. 
Subsequent studies incorporated visual grounding, first by aligning words with object categories~\citep{roy2002learning,yu2005emergence,xu2007word,yu2007unified,yu2013grounded}, and later by mapping words to richer visual features~\citep{qu2010context,jiayuan2019neuro,mao2021grammarbased,pratt2020grounded}. 
More recently, large-scale VLMs trained with paired text–image supervision have advanced grounding to finer levels of granularity, achieving region-level~\citep{li2022grounded,ma2023world,chen2023shikra,you2024ferret,wang2024cogvlm} and pixel-level~\citep{xia2023gsva,rasheed2024glamm,zhang2024groundhog} grounding, with strong performance on referring expression comprehension~\citep{chen2024revisiting}.

Recent work suggests that grounding emerges as a property of VLMs trained without explicit supervision, with evidence drawn from attention-based spatial localization~\citep{cao2024emerging,bousselham2024grounding} and cross-modal geometric correspondences~\citep{schnaus2025s}. 
However, all prior work focused exclusively on static final-stage models, overlooking the training trajectory, a crucial aspect for understanding when and how grounding emerges.
In addition, existing work has framed grounding through correlations between visual and textual signals, diverging from the definition by \citet{harnad1990symbol}, which emphasizes causal links from symbols to meanings.
To address these issues, we systematically examine learning dynamics throughout model training, applying causal interventions to probe model internals and introducing control groups to enable rigorous comparison.
% \freda{@Martin, see this new discussion?}
% \martin{I think causal needs to be defined through Mech Interp, in the current discussions, learning dynamics and control group do not guarantee causality}
% However, these findings depart from the classic grounding problem: they demonstrate correlational regularities in representations, not word-referent mapping, and therefore dilute the term by equating structural emergence with genuine grounding. \freda{What do you mean by structural emergence here?}
% \Freda{I'm thinking of criticizing the recent ``emergent grounding'' papers by some soft version of the following: what they are investigating is the (even non-gradient-based) attention maps, and the approach is fundamentally confounded.}

\boldstart{Emergent capabilities and learning dynamics of LMs.}
A central debate concerns whether larger language models exhibit genuinely new behaviors: \citet{wei2022emergent} highlight abrupt improvements in tasks, whereas later studies argue such effects are artifacts of thresholds or in-context learning dynamics~\citep{schaeffer2023emergent,lu2023emergent}. 
Beyond end performance, developmental analyses show that models acquire linguistic abilities in systematic though heterogeneous orders with variability across runs and checkpoints~\citep{sellam2021multiberts,blevins2022analyzing,biderman2023pythia,xia2023training,van2025polypythias}. 
Psychology-inspired perspectives further emphasize controlled experimentation to assess these behaviors~\citep{hagendorff2023machine}, and comparative studies reveal both parallels and divergences between machine and human language learning \citep{chang2022word,evanson2023language,chang2024characterizing,ma2025babysit}. 
At a finer granularity, hidden-loss analyses identify phase-like transitions~\citep{kangaslahti2025hidden}, while distributional studies attribute emergence to stochastic differences across training seeds \citep{zhao2025distributional}. 
Together, emergent abilities are not sharp discontinuities but probabilistic outcomes of developmental learning dynamics.
Following this line, we present a probability- and model internals--based analysis of how symbol grounding emerges during language model training.

\boldstart{Mechanistic interpretability of LMs.}
Mechanistic interpretability has largely focused on attention heads in Transformers~\citep{elhage2021mathematical, olsson2022context, meng2022locating, bietti2023birthtransformermemoryviewpoint, lieberum2023, wu2024retrieval}. 
A central line of work established that \textit{induction heads} emerge to support in-context learning~\citep[ICL;][]{elhage2021mathematical, olsson2022context}, with follow-up studies tracing their training dynamics~\citep{bietti2023birthtransformermemoryviewpoint} and mapping factual recall circuits~\citep{meng2022locating}. 
At larger scales, \citet{lieberum2023} identified specialized \textit{content-gatherer} and \textit{correct-letter} heads, and \citet{wu2024retrieval} showed that a sparse set of \textit{retrieval heads} is critical for reasoning and long-context performance. 
Relatedly, \citet{wang2023label} demonstrated that label words in demonstrations act as \textit{anchors}: early layers gather semantic information into these tokens, which later guide prediction. 
Based on these insights, \citet{bick2025understanding} proposed that retrieval is implemented through a coordinated \textit{gather-and-aggregate (G\&A)} mechanism: some heads collect content from relevant tokens, while others aggregate it for prediction. 
% They further showed that G\&A dynamics arise in both Transformers and SSMs, underscoring their architectural generality. 
Other studies extended this line of work by analyzing failure modes and training dynamics~\citep{wiegreffe2025} and contrasting retrieval mechanisms in Transformers and SSMs~\citep{ssms_retrieval}. 
% These findings motivate our focus on the attention head mechanism for symbol grounding. 
Whereas prior analyses typically investigate ICL with repeated syntactic or symbolic formats, our setup requires referential alignment between linguistic forms and their environmental contexts, providing a complementary testbed for naturalistic language grounding. 
% \vspace{-2pt}
\section{Method}
\label{sec:method}
% \vspace*{-8pt}

\subsection{Dataset and Tokenization}
\label{subsec:dataset}
% \vspace{-2pt}

% Our guiding principle is that symbol grounding should be learned from multimodal interactions, where environmental and conversational contexts offer the ground.
To capture the emergent grounding in multimodal interactions, we design a minimal testbed with a custom word-level tokenizer, in which every lexical item is represented in two corresponding forms: one token that appears in non-verbal descriptions (e.g., a \textit{book} in the scene description) and another that appears in utterances (e.g., \textit{book} in speech). 
We refer to these by environmental (\env) and linguistic tokens (\lan), respectively.
For instance, \textit{book}$_\env$ and \textit{book}$_\lan$ receive different integer indices from the tokenizer; that is, tokenization provides no explicit signal that these tokens are related, so any correspondence between them must be learned during training rather than inherited from surface forms.
Although our minimal testbed uses child-directed speech data, the framework is readily extensible to visual dialogue datasets, enabling the investigation of VLMs in realistic settings.
% We instantiate this framework with two datasets: child-directed speech transcripts and image-based dialogue.

\boldstart{Child-directed speech.}
The Child Language Data Exchange System \citep[CHILDES; ][]{macwhinney2000childes} provides transcripts enriched with environmental annotations.\footnote{See the manual for data usage: \url{https://talkbank.org/0info/manuals/CHAT.pdf}} 
We use the spoken utterances as the linguistic tokens (\lan) and the environmental descriptions as the environment tokens (\env). 
The environmental context is drawn from three annotation types:
\begin{itemize}[topsep=-4pt,itemsep=0pt]
    \item \textbf{Local events}: simple events, pauses, long events, or remarks interleaved with the transcripts.
    \item \textbf{Action tiers}: actions performed by the speaker or listener (e.g., \texttt{\%act: runs to toy box}). 
    These also include cases where an action replaces speech (e.g., \texttt{0 [\% kicks the ball]}).
    \item \textbf{Situational tiers}: situational information tied to utterances or to larger contexts (e.g., \texttt{\%sit: dog is barking}).
\end{itemize}

\boldstart{Image-grounded dialogue.}
To move beyond the minimal textual proxies, we create an image-grounded dialogue setup using the Visual Dialog dataset~\citep{das2017visual}, which pairs MSCOCO images~\citep{lin2014microsoft} with sequential multi-turn question-answering dialogues that exchange information about each image.
% \sout{using the same dataset as the caption-grounded dialogue setting.} 
% \jycc{the same dataset as the caption-grounded dialogue? the only difference one has the environmental tokens from caption, the other from images? }\freda{yes.}
Here, a frozen vision transformer ~\citep[ViT; ][]{dosovitskiy2020image} converts an RGB image into patch embeddings, with each embedding treated as an \env token, analogously to the visual tokens in modern VLMs.\footnote{
~Note that these ``visual tokens'' are continuously valued and do not correspond to discrete symbol-like tokens. They, as a whole, can be considered tensor representations of an image.
}
% Here, the environmental tokens are derived directly from RGB images: each image is tokenized into patch embeddings using a frozen vision transformer ~\citep[ViT; ][]{dosovitskiy2020image}, and each patch embedding is treated as an \env token, analogously to the visual tokens in modern VLMs. 
We use DINOv2~\citep{oquab2024dinov2} as our ViT tokenizer---because it is trained exclusively on vision data without auxiliary text supervision~\citep[unlike CLIP;][]{radford2021learning}, it ensures that environmental tokens encode strictly visual information.
The dialogues correspond to linguistic tokens (\lan), forming realistic multimodal interactions where conversational utterances are grounded directly in visual input.
% We use the DINOv2~\citep{oquab2024dinov2} as our ViT tokenizer, as it is trained purely on vision data without auxiliary text supervision~\citep[in contrast to models like CLIP;][]{radford2021learning}, thereby ensuring that environmental tokens capture only visual information.
% \sout{The linguistic tokens (\lan) remain unchanged from the caption-grounded dialogue setting,} % This setup allows us to evaluate whether the grounding effects observed in controlled text-based contexts also extend to real multimodal settings. 
% \freda{I feel the original last sentence is not necessary---we haven't said anything about the results.}
% \martin{Okie}
%\jycc{Table 1, a figure may be better than a table, will take less space too.}

We also introduce an intermediate setup: caption-grounded visual dialogue, which uses MSCOCO image captions as the grounding context. 
This configuration bridges the gap between the synthetic child-directed speech environment and the realistic image-grounded dialogue settings. 
Further details can be found in Appendix~\ref{sec:Caption-grounded}.

% \vspace{-2pt}
\subsection{Evaluation Protocol}
\label{subsec:eval}
% \vspace{-2pt}
We assess symbol grounding with a contrastive test that asks whether a model assigns a higher probability to the correct linguistic token when the matching environmental token is in context, following the idea of priming in psychology. 
% \freda{revisit}
This evaluation applies uniformly across datasets (Table~\ref{tab:data}): in CHILDES, environmental priming comes from descriptive contexts; in image-grounded dialogue, from ViT-derived visual tokens.
We compare the following conditions:

\begin{table*}[!t]
    \caption{Training and test examples across datasets with target word \textcolor{blue}{\it book}. The training examples combine environmental tokens (\colorbox{tticblue!10}{\env; shaded}) with linguistic tokens (\lan). Test examples are constructed with either matched (\textcolor{blue}{\it book}) or mismatched (\textcolor{red}{\it toy}) environmental contexts, paired with corresponding linguistic prompts. 
    Note that in child-directed speech, \textcolor{blue}{\it book$_\env$} and \textcolor{blue}{\it book$_\lan$} are two distinct tokens received by LMs.
    % Child-directed speech, caption- and image-grounded dialogue illustrate how grounding relations between \env and \lan tokens are instantiated across modalities. 
    % \textcolor{red}{\textbf{Owen: can we emphasize ENV to avoid confusion?}} 
    }
    \vspace{-3pt}
    \label{tab:data}
    \centering
    \scalebox{0.9}{
    \begingroup
    \setlength{\tabcolsep}{5.5pt}
    \hspace*{-7pt}
    \begin{tabular}{m{0.10\linewidth} m{0.17\linewidth} m{0.23\linewidth} m{0.15\linewidth} m{0.18\linewidth} m{0.11\linewidth}}
    \toprule
    \multirow{2}{*}{\parbox{20pt}{\bf Dataset}} & \multicolumn{2}{c}{\bf Training Example} & \multicolumn{3}{c}{\bf Test Example} \\
    \cmidrule(lr){2-3}\cmidrule(lr){4-6}
     &
    \multicolumn{1}{c}{\cellcolor{tticblue!10} \bf \env} &
    \multicolumn{1}{c}{\bf \lan} &
    \multicolumn{1}{c}{\cellcolor{tticblue!10} \bf \env Match} &
    \multicolumn{1}{c}{\cellcolor{tticblue!10} \bf \env Mismatch} &
    \multicolumn{1}{c}{\bf \lan} \\
    \midrule
    {\bf Child-Directed Speech} &
    \cellcolor{tticblue!10} \textit{$\langle$CHI$\rangle$ takes \textcolor{blue}{book} from mother} &
    \textit{$\langle$CHI$\rangle$ what's that $\langle$MOT$\rangle$ a \textcolor{blue}{book} in it ...} &
    \cellcolor{tticblue!10} \textit{$\langle$CHI$\rangle$ asked for a new \textcolor{blue}{book}} &
    \cellcolor{tticblue!10} \textit{$\langle$CHI$\rangle$ asked for a new \textcolor{red}{toy}} &
    \textit{$\langle$CHI$\rangle$ I love this \underline{\hspace{3em}}} \\
    % \midrule
    % {\bf Caption-Grounded Dialogue} &
    % \cellcolor{tticblue!10} \textit{a dog appears to be reading a \textcolor{blue}{book} with a full bookshelf behind} &
    % \textit{$\langle$Q$\rangle$ can you tell what \textcolor{blue}{book} it's reading $\langle$A$\rangle$ the marriage of true minds by stephen evans} &
    % \cellcolor{tticblue!10} \textit{this is a \textcolor{blue}{book}} &
    % \cellcolor{tticblue!10} \textit{this is a \textcolor{red}{toy}} &
    % \textit{$\langle$Q$\rangle$ can you name this object $\langle$A$\rangle$ \underline{\hspace{3em}}} \\
    \midrule
    {\bf Image-Grounded Dialogue} &
    \cellcolor{tticblue!10} 
    \begin{center}
    \vspace{-5pt}\includegraphics[height=40pt]{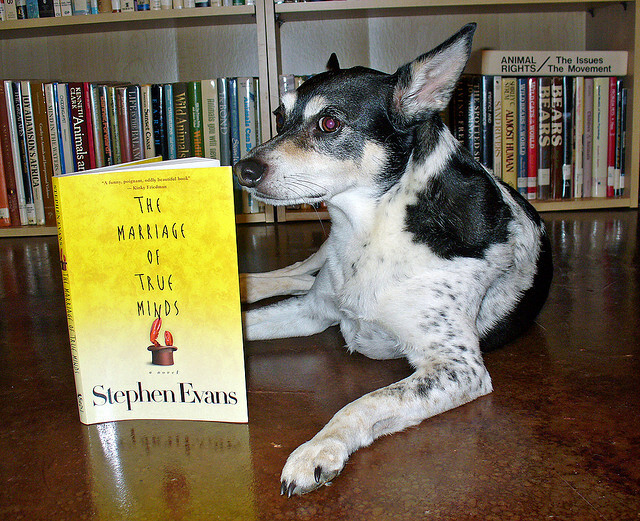}\vspace{-10pt}
    \end{center} &
    \textit{$\langle$Q$\rangle$ can you tell what \textcolor{blue}{book} it's reading $\langle$A$\rangle$ the marriage of true minds by stephen evans} &
    \cellcolor{tticblue!10} 
    \begin{center}
    \vspace{-5pt}\includegraphics[height=40pt]{figs/data/book-test.jpg}\vspace{-10pt}
    \end{center} &
    \cellcolor{tticblue!10} 
    \begin{center}
    \vspace{-5pt}\includegraphics[height=40pt]{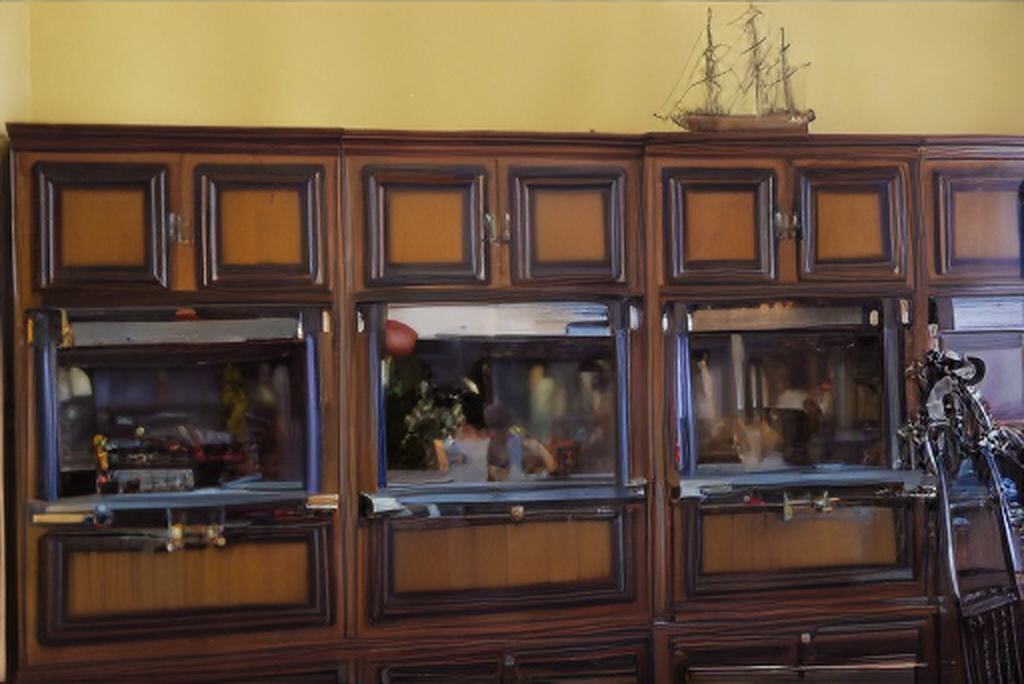}\vspace{-10pt}
    \end{center} &
    \textit{what do we have here? \underline{\hspace{3em}}} \\
    \bottomrule
    \end{tabular}
    \endgroup}
    \vspace*{-10pt}
\end{table*}

\begin{itemize}[topsep=-3pt,itemsep=0pt]
    \item \textbf{Match (experimental condition)}: 
    The context contains the corresponding \env token for the target word, and the model is expected to predict its \lan counterpart.
    \item \textbf{Mismatch (control condition)}: 
    The context is replaced with a different \env token. 
    The model remains tasked with predicting the same \lan token; however, in the absence of corresponding environmental cues, its performance is expected to be no better than chance.
\end{itemize}
% \jycc{it may help reading by defining the conditions first, and then specifying the hypothesis/expection as in both conditions, surprisal is measured etc...}

For example (first row in Table~\ref{tab:data}), when evaluating the word $\textit{book}_\lan$, the input context is 
\begin{align}
    \vspace{-2pt}
    &\langle\textit{CHI}\rangle \textit{ asked}_\env \textit{ for}_\env \textit{ a}_\env \textit{ new}_\env \textit{ book}_\env \textit{ } \nonumber \\
    & \langle\textit{CHI}\rangle \textit{ I}_\lan \textit{ love}_\lan \textit{ this}_\lan \textit{ } \underline{\hspace{3em}}, 
    \vspace{-2pt}
    \label{eq:example}
\end{align}
where the model is expected to predict $\textit{book}_\lan$ for the blank, and the role token \textit{$\langle$CHI$\rangle$} indicates the involved speaker or actor's role being a child.
In the control (mismatch) condition, the environmental token \textit{box}$_\env$ is replaced by another valid noun such as \textit{toy}$_\env$.

% For example, when evaluating the word \texttt{box}$_\lan$, the input sequence is:\freda{Is it possible to have an example appear in a separate line?}
% \texttt{\texttt{$\langle$CHI$\rangle$} jumped$_\env$ beside$_\env$ a$_\env$ large$_\env$ box$_\env$ \texttt{$\langle$CHI$\rangle$} I$_\lan$ found$_\lan$ a$_\lan$ cool$_\lan$ \underline{\hspace{3em}}},
% where the model is expected to predict \texttt{box}$_\lan$. 

\boldstart{Context templates.} 
For a target word $v$ with linguistic token $v_\lan$ and environmental token $v_\env$, we denote $\overline{C}_v$ as a set of context templates of $v$. 
For example, when $v=\textit{book}$, a $\overline{c} \in \overline{C}_v$ can be 
\begin{align}
    \vspace{-2pt}
    &\langle\textit{CHI}\rangle \textit{ asked}_\env \textit{ for}_\env \textit{ a}_\env \textit{ new}_\env \textit{ } \texttt{[FILLER]} \nonumber \\
    &\langle\textit{CHI}\rangle \textit{ I}_\lan \textit{ love}_\lan \underline{\hspace{3em}},
    \label{eq:template}
    \vspace{-2pt}
\end{align}
where \texttt{[FILLER]} is to be replaced with an environmental token, and the blank indicates the expected prediction as in Eq.~\eqref{eq:example}.
In the match condition, the context $\overline{c}(v)$ is constructed by replacing \texttt{[FILLER]} with $v_\env$ in $\overline{c}$.
In the mismatch condition, the context $\overline{c}(u)$ uses $u_\env (u\neq v)$ as the filler, while the prediction target remains $v_\lan$.

For the choices of $v$ and $u$, we construct the vocabulary $V$ with 100 nouns from the MacArthur--Bates Communicative Development Inventories~\citep{fenson2007macarthur} that occur frequently in our corpus. 
Each word serves once as the target, with the remaining $M=99$ used to construct mismatched conditions.
For each word, we create $N=10$ context templates, which contain both \env and \lan tokens.
Details of the vocabulary and context template construction can be found in the Appendix~\ref{app:data}. 
% \freda{Appendix pointer here} \freda{Note that we haven't defined the LM acronym yet}

\boldstart{Grounding information gain.}
Following prior work, we evaluate how well an LM learns a word using the mean surprisal over instances. 
The surprisal of a word $w$ given a context $c$ is defined as  
$
s_{\boldsymbol{\theta}}(w \mid c) = -\log P_{\boldsymbol{\theta}}(w \mid c),
$  
where $P_{\boldsymbol{\theta}}(w \mid c)$ denotes the probability, under an LM parameterized by ${\boldsymbol{\theta}}$, that the next word is $w$ conditioned on the context $c$.
Here, $s_{\boldsymbol{\theta}}(w \mid c)$ quantifies the unexpectedness of predicting $w$, or the pointwise information carried by $w$ conditioned on the context.
%\freda{A note here to keep in mind: we haven't emphasized the autoregressiveness yet in the previous context. }
% In our setting, $w = v_\lan$ and $c$ specifies either the correct ground $v_\env$ (match) or a mismatched ground $u_\env$ with $u \neq v$.
% For each target word $v$, we construct $N$ context templates $\{c_1, \dots, c_N\}$. 
% For each template $c_n$, we evaluate the model under the match condition and under $M$ mismatched conditions. 
% \jycc{the equation is missing index across M for $u$ env}

The \textit{grounding information gain} $G_{\boldsymbol{\theta}}(v)$ for $v$ is defined as
\begin{align*}
G_{\boldsymbol{\theta}}(v)  = \frac{1}{N} \sum_{n=1}^N
       \Biggl( \frac{1}{M} \sum_{u \neq v}^M
     \Big[ & s_{\boldsymbol{\theta}}\left(v_\lan \mid \overline{c}_n\left(u_\env\right)\right) \\
         - & s_{\boldsymbol{\theta}}\left(v_\lan \mid \overline{c}_n\left(v_\env\right)\right) \Big] \Biggr).
\end{align*}
This is a sample-based estimation of the expected log-likelihood ratio between the match and mismatch conditions
\vspace{-5pt}
\begin{align*}
G_{\boldsymbol{\theta}}(v) = \mathbb{E}_{c,u}\left[ \log \frac{P_{\boldsymbol{\theta}}(v_\lan\mid c, v_\env)}{P_{\boldsymbol{\theta}}(v_\lan\mid c, u_\env)} \right],
\end{align*}
\vspace{-5pt}

which quantifies how much more information the matched ground provides for predicting the linguistic form, compared to a mismatched one.
% \freda{@Martin, can you see if this is correct?}
% which quantifies how much evidence the correct environmental ground provides for the hypothesis that the observed linguistic form was generated from it, or equivalently, the average number of bits by which uncertainty is reduced when conditioning on the correct ground rather than an incorrect one.
A positive $G_{\boldsymbol{\theta}}(v)$ indicates that the matched environmental token increases the predictability of its linguistic form.
We report $G_{\boldsymbol{\theta}} = \frac{1}{|V|} \sum_{v \in V} G_{\boldsymbol{\theta}}(v)$, and track $G_{{\boldsymbol{\theta}}^{(t)}}$ across training steps $t$ to analyze how grounding emerges over time.

\subsection{Model Training}

We train LMs from random initialization, ensuring that no prior linguistic knowledge influences the results. 
Our training uses the standard causal language modeling objective, as in most generative LMs. 
To account for variability, we repeat all experiments with 5 random seeds, randomizing both model initialization and corpus shuffle order. 
Our primary architecture is Transformer~\citep{vaswani2017attention} in the style of GPT-2 \citep{radford2019language} with 18, 12, and 4 layers, with all of them having residual connections.
We extend the experiments to 4-layer unidirectional LSTMs~\citep{hochreiter1997long} with no residual connections, as well as 12- and 4-layer state-space models~\citep[specifically, Mamba-2; ][]{dao2024transformers}.
For fair comparison with LSTMs, the 4-layer Mamba-2 models do not involve residual connections, whereas the 12-layer ones do.
For multimodal settings, while standard LLaVA~\citep{liu2023visual} uses a two-layer perceptron to project ViT embeddings into the language model, we bypass this projection in our case and directly feed the DINOv2 representations into the LM.
We obtain the developmental trajectory of the model by saving checkpoints at various training steps, sampling more heavily from earlier steps, following \citet{chang2022word}. Notice that datasets used to pretrain these LMs are the original version of those mentioned in section~\ref{subsec:dataset}, and the context templates mentioned in section~\ref{subsec:eval} are strictly inference-only and are never seen during model training.
% \jycc{what's the significance of these 33 checkpoints. why these are good representation? add one sentence here.} 
% \freda{TODO: Appendix to detail the step sampling strategy.}
% \martin{Models: 18 trans, 12 trans, 4 trans, 12 mamba, 4 mamba, 4 lstm, all 4-layers have no res}

\section{Behavioral Evidence}
\label{sec:exp_behave}

\begin{figure*}[!t]
    \centering
    \begin{subfigure}[t]{0.235\textwidth}
        \centering
        \includegraphics[width=1.05\columnwidth]{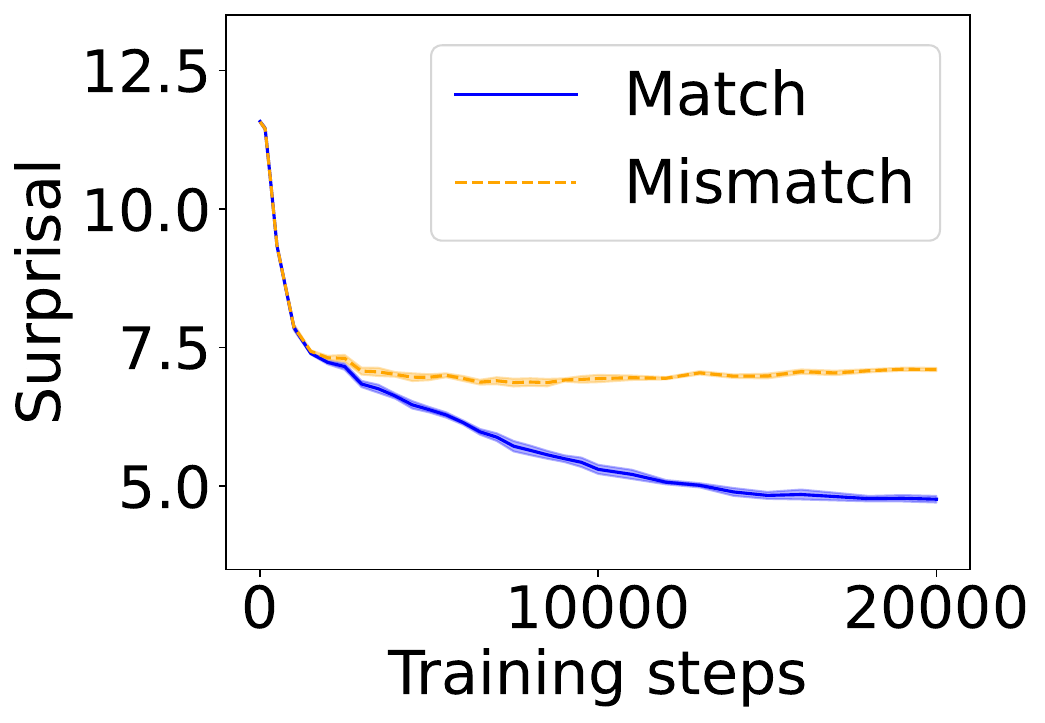}
        \vspace*{-15pt}
        \caption{12-layer Transformer.}
        \label{fig:sps-childes-transformer}
    \end{subfigure}
    ~
    \begin{subfigure}[t]{0.235\textwidth}
        \centering
        \includegraphics[width=1.05\columnwidth]{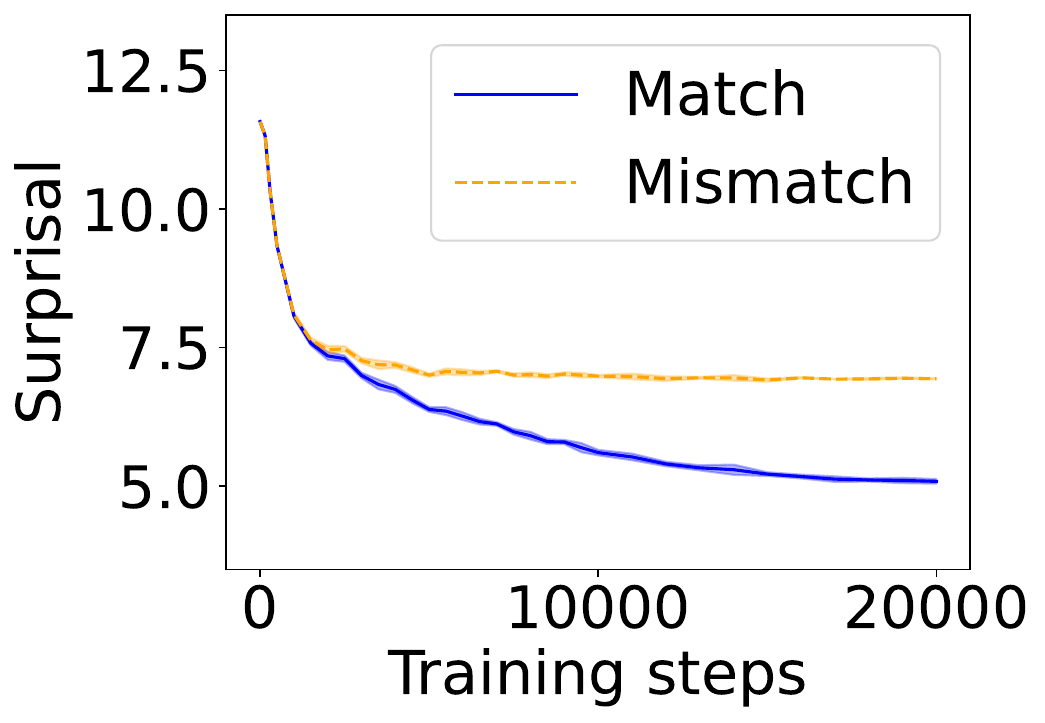}
        \vspace*{-15pt}
        \caption{4-layer Transformer.}
        \label{fig:sps-childes-transformer-4layer}
    \end{subfigure}
    ~
    \begin{subfigure}[t]{0.235\textwidth}
        \centering
        \includegraphics[width=1.05\columnwidth]{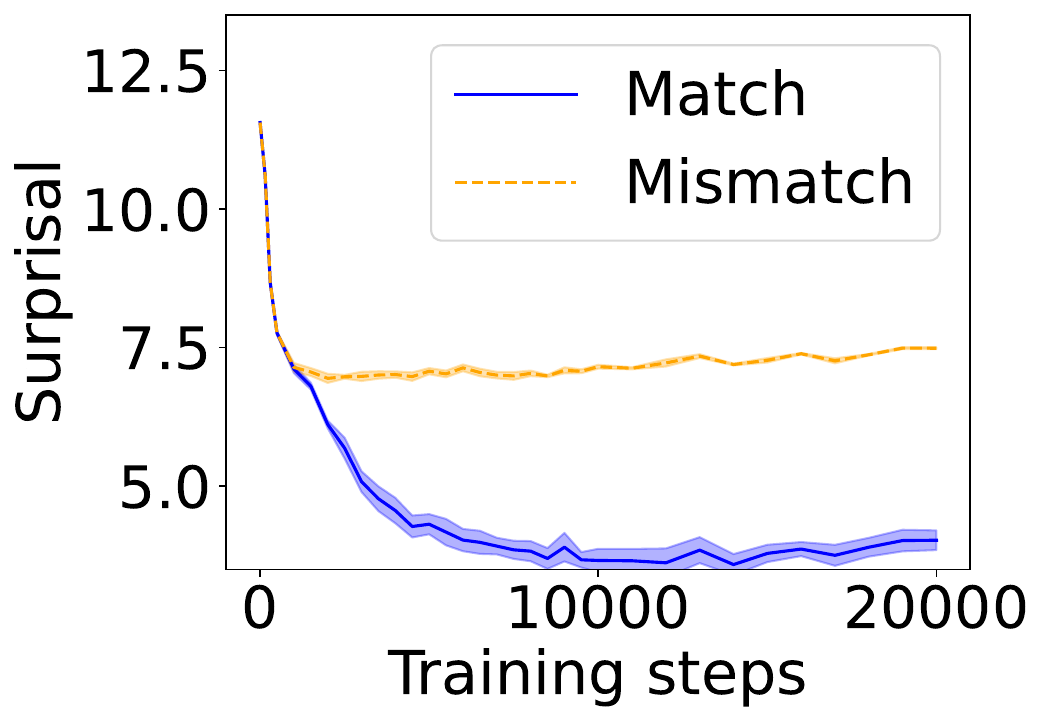}
        \vspace*{-15pt}
        \caption{4-layer Mamba 2.}
        \label{fig:sps-childes-mamba-4layer}
    \end{subfigure}
    ~
    \begin{subfigure}[t]{0.235\textwidth}
        \centering
        \includegraphics[width=1.05\columnwidth]{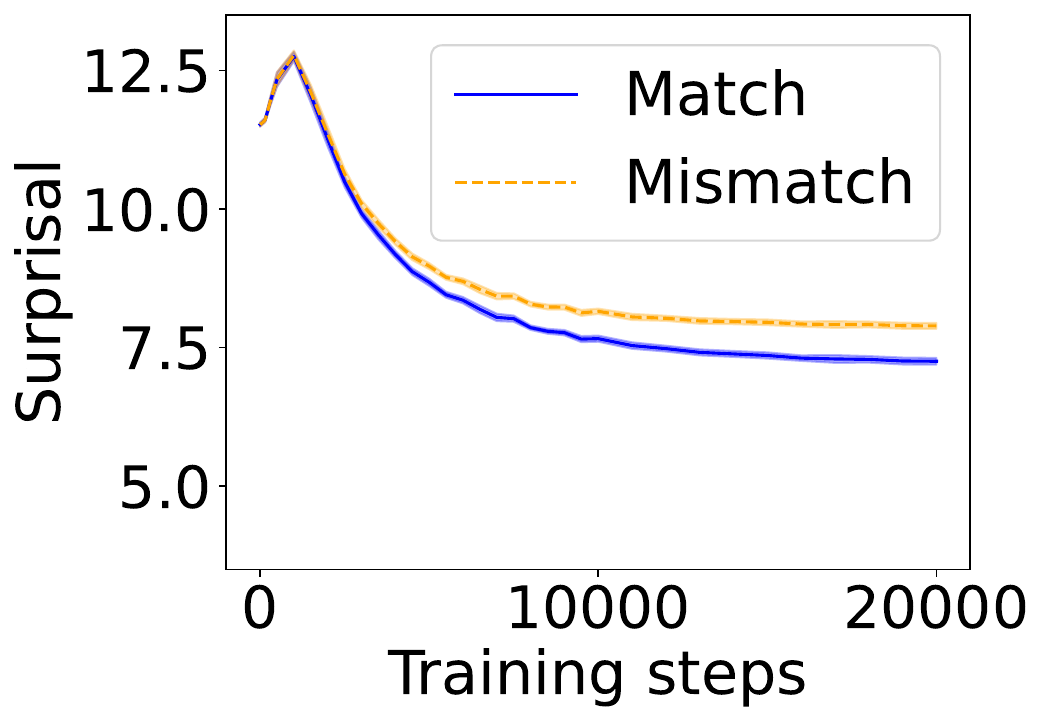}
        \vspace*{-15pt}
        \caption{4-layer LSTM.}
        \label{fig:sps-childes-lstm}
    \end{subfigure}
    \vspace*{-5pt}
    \caption{Average surprisal of the experimental and control conditions over training steps. \vspace*{-10pt}}
    \label{fig:surprisal-childes}
\end{figure*}

\begin{figure*}[!t]
    \centering
    \begin{subfigure}[t]{.235\textwidth}
        \centering
        \includegraphics[width=1.05\columnwidth]{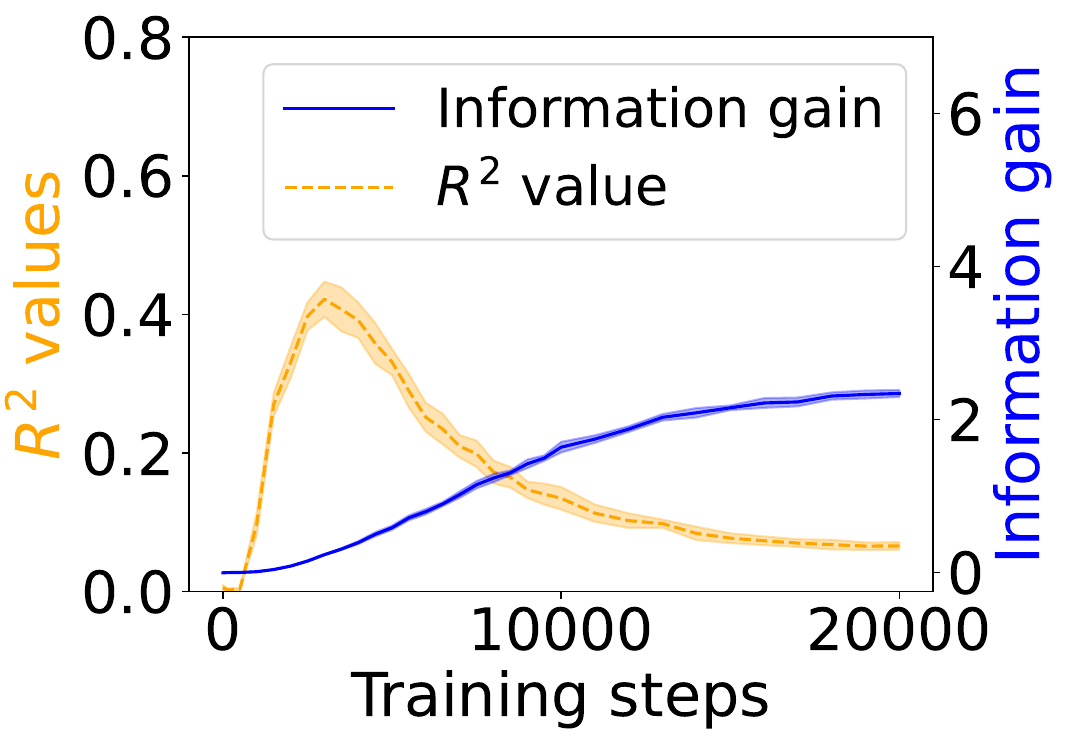}
        \vspace*{-15pt}
        \caption{12-layer Transformer.}
        \label{fig:r2-childes-transformer}
    \end{subfigure}
    ~
    \begin{subfigure}[t]{.235\textwidth}
        \centering
        \includegraphics[width=1.05\columnwidth]{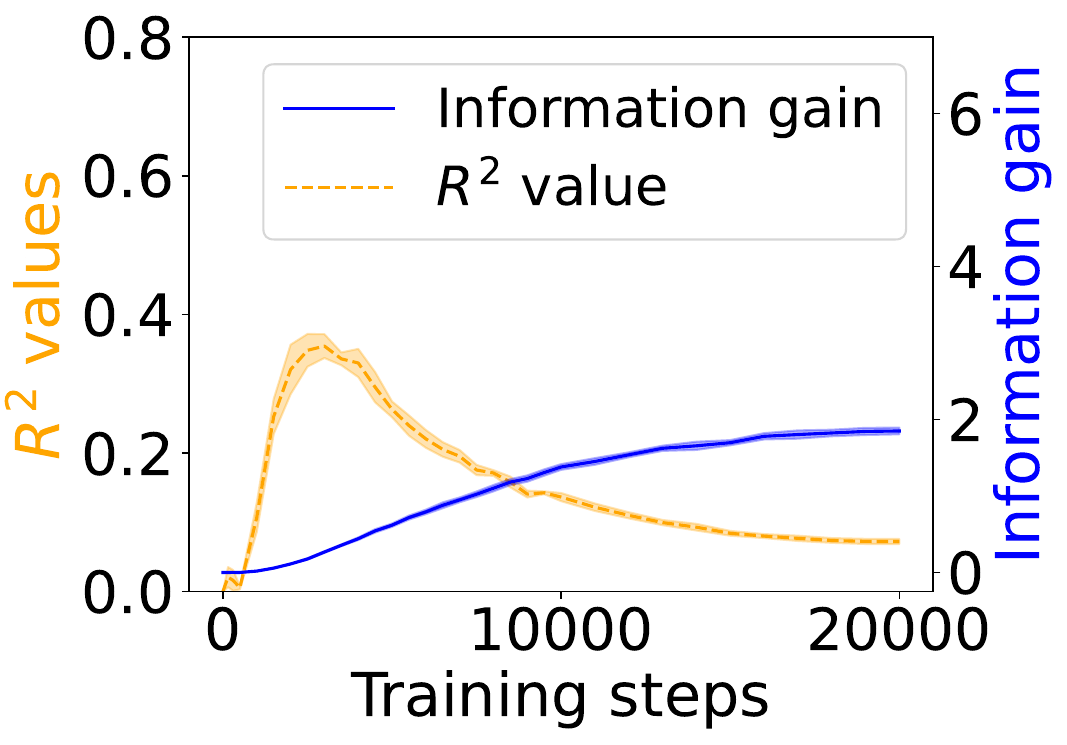}
        \vspace*{-15pt}
        \caption{4-layer Transformer.}
        \label{fig:r2-childes-transformer-4layer}
    \end{subfigure}
    ~
    \begin{subfigure}[t]{.235\textwidth}
        \centering
        \includegraphics[width=1.05\columnwidth]{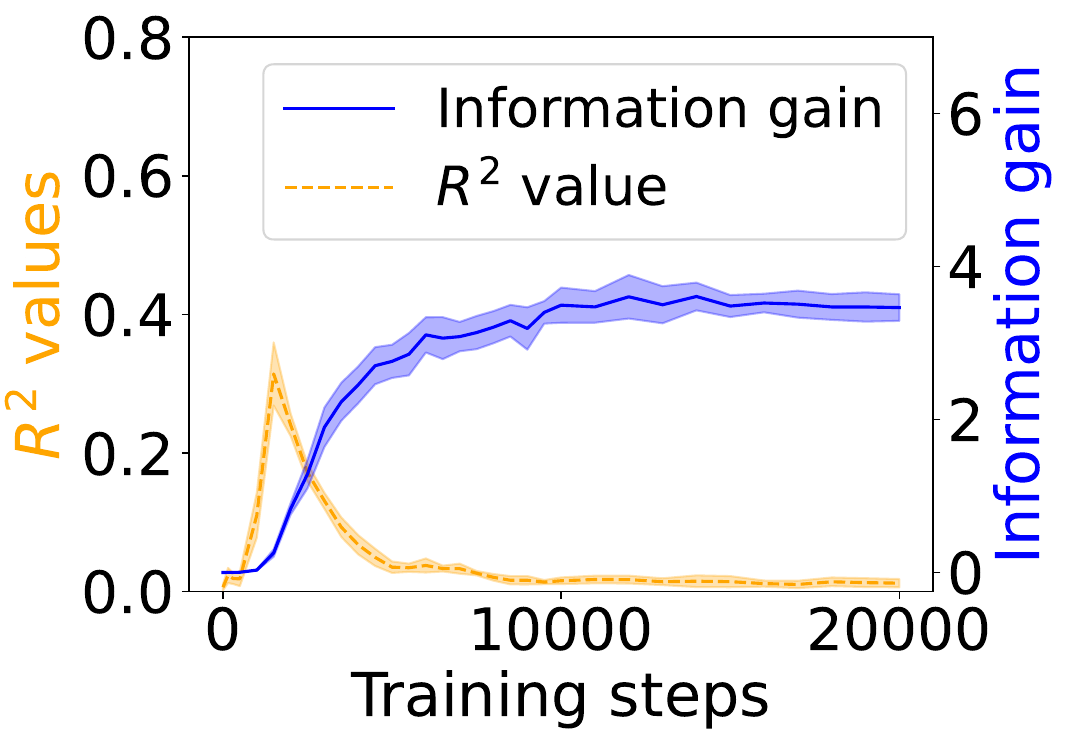}
        \vspace*{-15pt}
        \caption{4-layer Mamba 2.}
        \label{fig:r2-childes-mamba-4layer}
    \end{subfigure}
    ~
    \begin{subfigure}[t]{.235\textwidth}
        \centering
        \includegraphics[width=1.05\columnwidth]{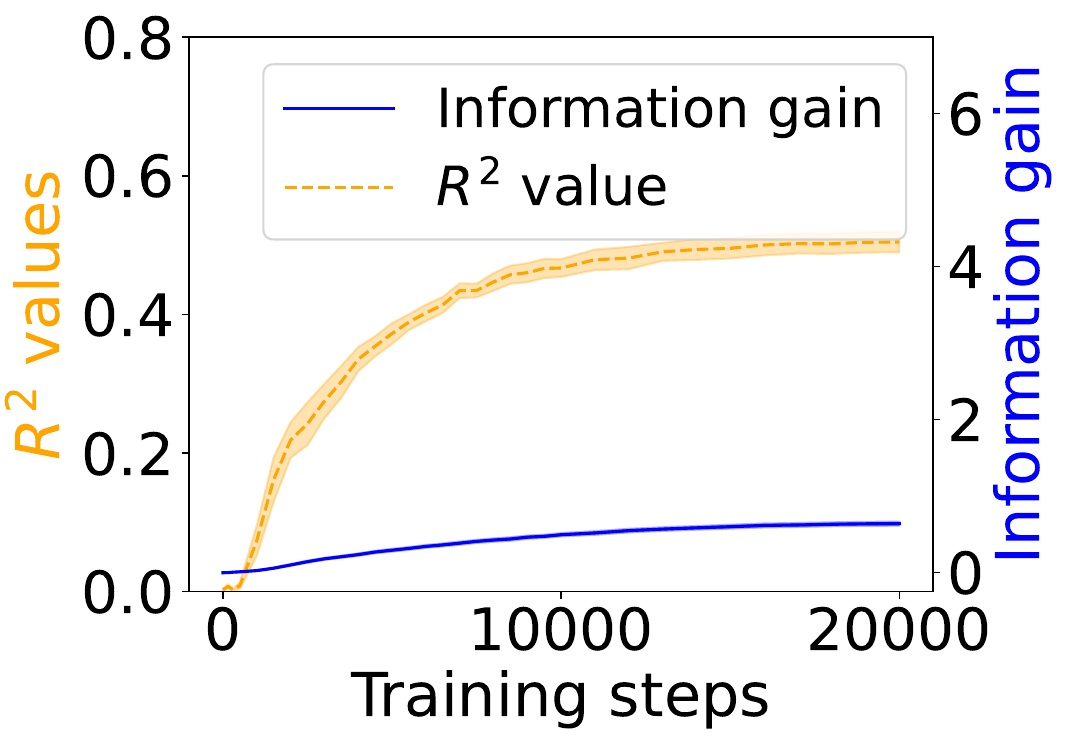}
        \vspace*{-15pt}
        \caption{4-layer LSTM.}
        \label{fig:r2-childes-lstm}
    \end{subfigure}
    \vspace*{-5pt}
    \caption{Grounding information gain and its correlation to the co-occurrence of linguistic and environment tokens over training steps. \vspace*{-5pt}}
    \label{fig:r2-childes}
\end{figure*}

\subsection{Behavioral Evidence of Emergent Grounding}
\label{subsec:behavioral}

In this section, we ask: \textbf{Does symbol grounding emerge behaviorally in our settings?} 
We first test whether models show systematic surprisal reduction when predicting a linguistic token when its environmental counterpart is in context (Figure~\ref{fig:surprisal-childes}, where the gap between the lines represents the grounding information gain). 
% Figure~\ref{fig:surprisal-childes} plots average surprisal in the match and mismatch conditions, with their gap representing the grounding information gain, evaluated on the CHILDES dataset.
For Transformers (Figures~\ref{fig:sps-childes-transformer} and \ref{fig:sps-childes-transformer-4layer}) and Mamba-2 (Figure~\ref{fig:sps-childes-mamba-4layer}), surprisal in the match condition decreases steadily while that in the mismatch condition enters a high-surprisal plateau early, indicating that the models leverage environmental context to predict the linguistic form. 
% A similar pattern appears in Mamba-2, although the 12-layer version shows sharper learning dynamics initially and drifts later, consistent with the tendency of Mamba to overfit local pattern shortcuts~\citep{you2025revealing}. 
In contrast, the unidirectional LSTM (Figure~\ref{fig:sps-childes-lstm}) shows little separation between the conditions, reflecting the absence of grounding.
Overall, these results provide behavioral evidence of emergent grounding: in sufficiently expressive architectures (Transformers and Mamba-2), the correct environmental context reliably lowers surprisal for its linguistic counterpart, whereas LSTMs fail to exhibit this effect, marking an architectural boundary on where grounding can emerge.

% \martin{Add 4/18 layer Transformer.}

\subsection{Behavioral Effects Beyond Co-Occurrence} 
% \freda{surpass $\rightarrow$ beyond?}

A natural concern is that the surprisal reductions might be fully explainable by shallow statistics: \textbf{the models might have simply memorized frequent co-occurrences of \env and \lan tokens, without learning a deeper and more general mapping.}
We test this hypothesis by comparing the tokens' co-occurrence with the grounding information gain in the child-directed speech data.

We define co-occurrence between the corresponding \env and \lan tokens at the granularity of a 512-token training chunk. 
For each target word $v$, we count the number of chunks in which both its \env and \lan tokens appear.
% \jycc{what is a record? utterance level? discourse level in CHILDS?}\freda{how about chunk?} \jycc{chunks or data instances?}
Following standard corpus-analysis practice, these raw counts are log-transformed.
For each model checkpoint, we run linear regression between the log co-occurrence and the grounding information gain of words, obtaining an $R^2$ statistic as a function of training time.

Figure~\ref{fig:r2-childes} shows the $R^2$ values (orange) alongside the grounding information gain (blue) for different architectures. 
In both the Transformer and Mamba-2, $R^2$ rises sharply at the early steps but then goes down, even if the grounding information gain continues increasing. 
These results suggest that grounding in Transformers and Mamba-2 cannot be fully accounted for by co-occurrence statistics: while models initially exploit surface co-occurrence regularities, later improvements in grounding diverge from these statistics, indicating reliance on richer and more complicated features acquired during training.
In contrast, LSTM shows persistently increasing $R^2$ but little increase in grounding information gain over training steps, suggesting that it encodes co-occurrence but lacks the architectural mechanism to transform it into predictive grounding.
% \freda{I grouped the discussions to (I feel) better streamline the logical flow.}

% \vspace{-5pt}
\subsection{Results on Image-Grounded Dialogue}
% \vspace{-5pt}

\begin{figure}[!t]
    \centering
    % \begin{subfigure}[t]{.21\textwidth}
    %     \centering
    %     \includegraphics[width=1.03\columnwidth]{figs/surprisal_overtime/visdial_textonly_transformer.pdf}
    %     \vspace*{-15pt}
    %     \caption{Surprisal curves (w/ caption).}
    %     \label{fig:sps-visdial-cap}
    % \end{subfigure}
    % ~
    \begin{subfigure}[t]{.224\textwidth}
        \centering
        \includegraphics[width=1.05\columnwidth]{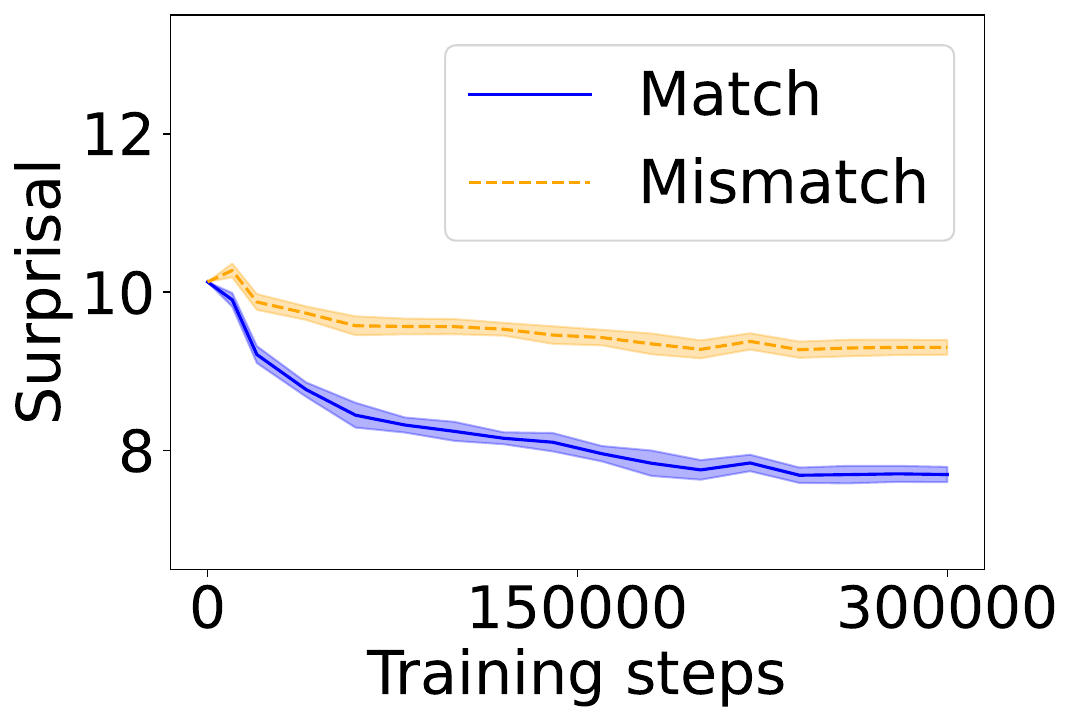}
        \vspace*{-15pt}
        \caption{Surprisal curves.}
        \label{fig:sps-visdial-img}
    \end{subfigure}
    ~
    % \begin{subfigure}[t]{.23\textwidth}
    %     \centering
    %     \includegraphics[width=1.04\columnwidth]{figs/coocurrence/visdial_textonly_transformer.pdf}
    %     \vspace*{-15pt}
    %     \caption{$R^2$ and information gain (w/ caption).}
    %     \label{fig:r2-visdial-cap}
    % \end{subfigure}
    % ~
    \begin{subfigure}[t]{.24\textwidth}
        \centering
        \includegraphics[width=1.05\columnwidth]{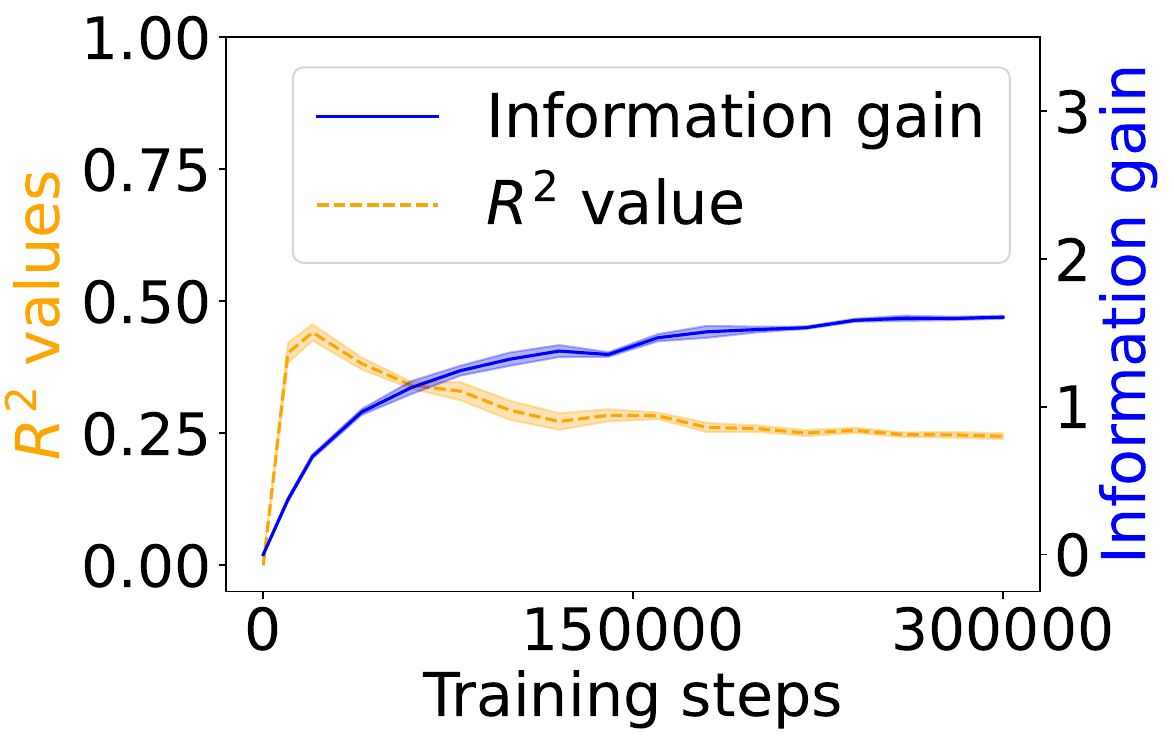}
        \vspace*{-15pt}
        \caption{$R^2$ and information gain.}
        \label{fig:r2-visdial-img}
    \end{subfigure}
    \vspace*{-5pt}
    \caption{Average surprisal of the experimental and control conditions, as well as the grounding information gain and its correlation to the co-occurrence of linguistic and environment tokens over training steps. 
    All results are from a 12-layer Transformer model on image-grounded dialogue.
        \vspace*{-10pt}
    }
    \label{fig:visdial}
\end{figure}

We next test whether the grounding effects observed in CHILDES generalize to VLMs, using the Visual Dialog dataset. 
In this setting, the environmental ground is supplied by image features (Table~\ref{tab:data}), and mismatched contexts are generated via image inpainting with Stable Diffusion 2~\citep{rombach2022high}, which regenerates the ground-truth mask region corresponding to the target referent.

We train 12-layer Transformers with 5 random seeds.
Similarly as Figures~\ref{fig:sps-childes-transformer}--\ref{fig:sps-childes-transformer-4layer} and Figures~\ref{fig:r2-childes-transformer}--\ref{fig:r2-childes-transformer-4layer}, when images serve as the environmental ground, Transformers show a clear surprisal gap between match and mismatch conditions (Figure~\ref{fig:sps-visdial-img}), with the grounding information gain increasing steadily while $R^2$ peaks early and declines (Figure~\ref{fig:r2-visdial-img}), although the observed effect is slightly less pronounced.
These results confirm that emergent grounding cannot be fully explained by co-occurrence statistics.
% We also note that the overall $R^2$ values are notably higher than that in CHILDES, reflecting that environmental grounds and linguistic forms co-occur more frequently in Visual Dialog as captions are written to describe the image, and dialogue often reuses those terms.
% These findings demonstrate that emergent grounding extends beyond text-only child–caregiver dialogue. 
% The Transformer is able to exploit both caption-based and image-based environmental grounds to facilitate linguistic prediction in Visual Dialog. 

Overall, our findings demonstrate that Transformers are able to exploit environmental grounds in various modalities to facilitate linguistic prediction. 
% The fact that surprisal reductions and Information Gain persist after controlling for co-occurrence shows that the model learns systematic ENV-to-LAN mappings, even in high-dimensional multimodal settings. 
The smaller but consistent gains in the image-grounded case suggest that while grounding from visual tokens is harder, the same architectural dynamics identified in textual testbeds still apply.

\section{Mechanistic Explanation}
\label{sec:exp_mech}

In this section, we provide a mechanistic and interpretable account of the previous observation. 
We first draw hypotheses from a 12-layer Transformer trained on CHILDES with 5 random seeds, and extend the experiments to image-grounded dialogue (Section~\ref{sec:mechanistic-vlm}).

\begin{figure}[!t]
\centering
    \vspace*{-3pt}
    \begin{subfigure}[t]{\linewidth}
        \centering
        \includegraphics[width=\linewidth]{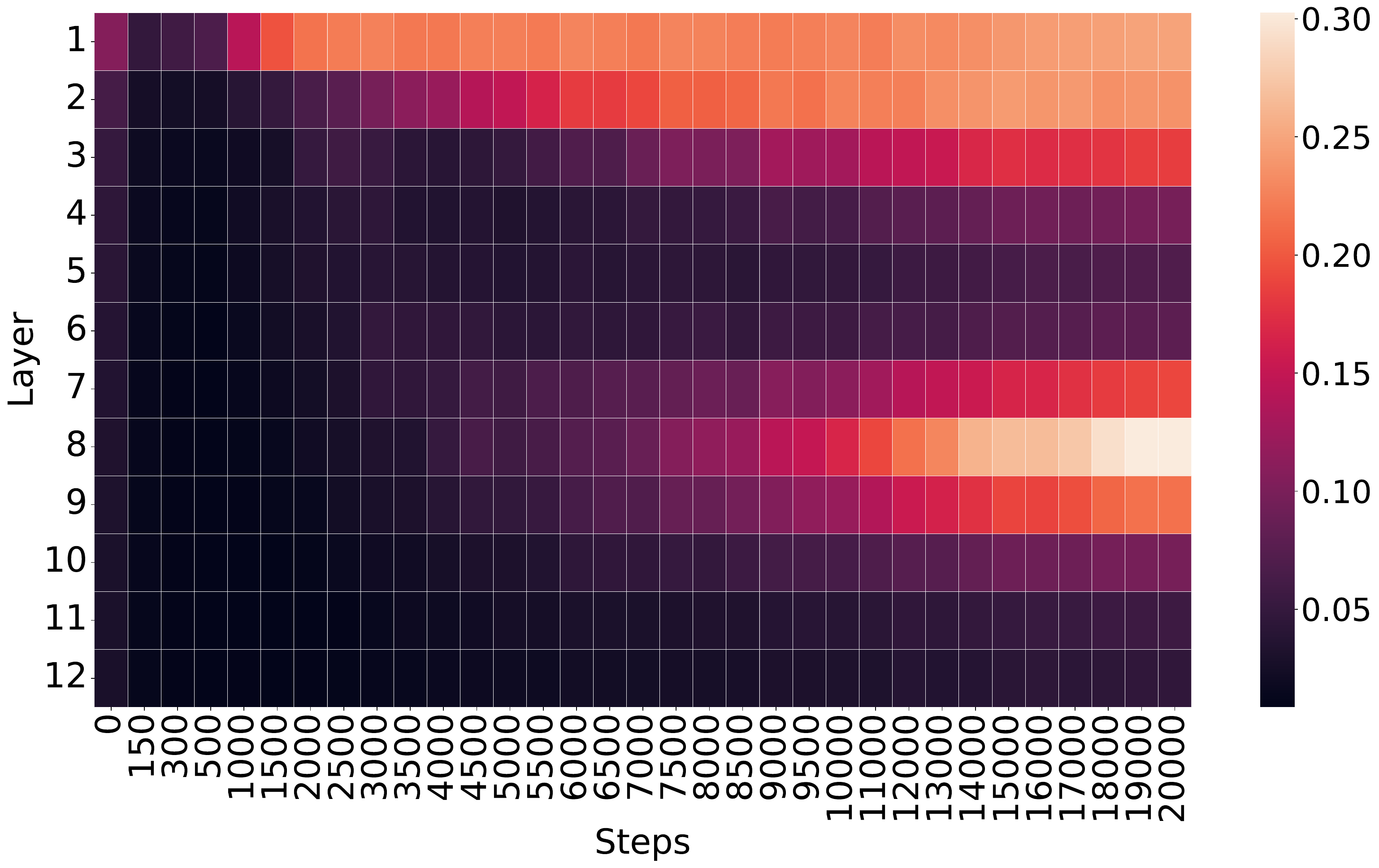}
        \vspace*{-15pt}
        \caption{Saliency of layer-wise attention from environmental to linguistic tokens across training steps.}
        \label{fig:saliency}
    \end{subfigure}
    ~
    \begin{subfigure}[t]{\linewidth}
        \centering
        \includegraphics[width=0.5\linewidth]{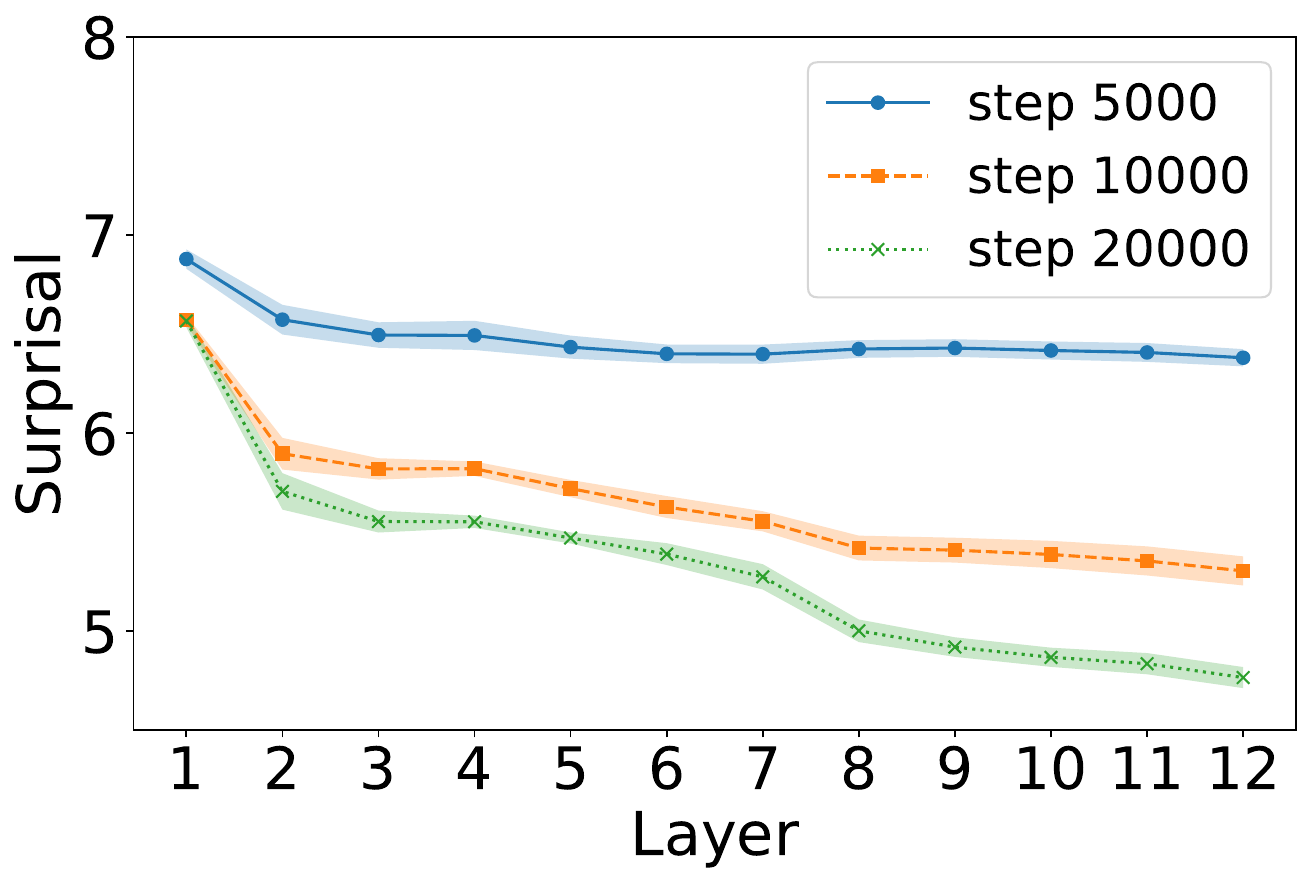}
        \vspace*{-5pt}
        \caption{Layer-wise tuned lens surprisals in the matched condition.}
        \label{fig:lens}
    \end{subfigure}
    \vspace*{-5pt}
    \caption{Mechanistic analysis of symbol grounding emergence. \vspace*{-10pt}}
    \label{fig:mechanistic}
    % \jyc{Figure 3b legend hard to read, although trends are there. may remind the reader the difference lighter vs. darker shades}
\end{figure}

% \vspace{-5pt}
\subsection{The Emergence of Symbol Grounding}
% \vspace{-5pt}

To provide a mechanistic account of symbol grounding, i.e., when it emerges during training and how it is represented in the network, we apply two interpretability analyses.

\boldstart{Saliency flow.} 
For each layer $\ell$, we compute a saliency matrix following~\citet{wang2023label}: $I_\ell = \left| \sum_h A_{h, \ell} \odot \frac{\partial \mathcal{L}}{\partial A_{h, \ell}} \right|$, 
where $A_{h,\ell}$ denotes the attention matrix of head $h$ in layer $\ell$. 
Each entry of $I_\ell$ quantifies the contribution of the corresponding attention weight to the cross-entropy loss $\mathcal{L}$, averaged across heads. 
Our analysis focuses on ground-to-symbol connections, i.e., flows from environmental ground (\env) tokens to the token immediately preceding (and predicting) their linguistic forms (\lan).

\boldstart{Probing with the Tuned Lens.}
We probe layer-wise representations using the Tuned Lens \citep{belrose2023eliciting}, which trains affine projectors to map intermediate hidden states to the final prediction space while keeping the LM output head frozen. 

\boldstart{Results.} 
Ground-to-symbol saliency is weak in the early stages of training but rises sharply later, peaking in layers 7–9 (Figure~\ref{fig:saliency}), suggesting that mid-layer attention may play a central role in establishing symbol–ground correspondences. 
In addition, Figure~\ref{fig:lens} shows that early layers remain poor predictors even at late training stages (e.g., after 20,000 steps), whereas surprisal begins to drop markedly from layer 7 at intermediate stages (step 10,000), suggesting a potential representational shift in the middle layers.

% \vspace{-5pt}
\subsection{Hypothesis: Gather-and-Aggregate Heads Implement Symbol Grounding}
% \vspace{-5pt}

\begin{figure}[!t]
    \centering

    \begin{subfigure}[t]{.48\linewidth}
        \centering
        \includegraphics[width=\linewidth]{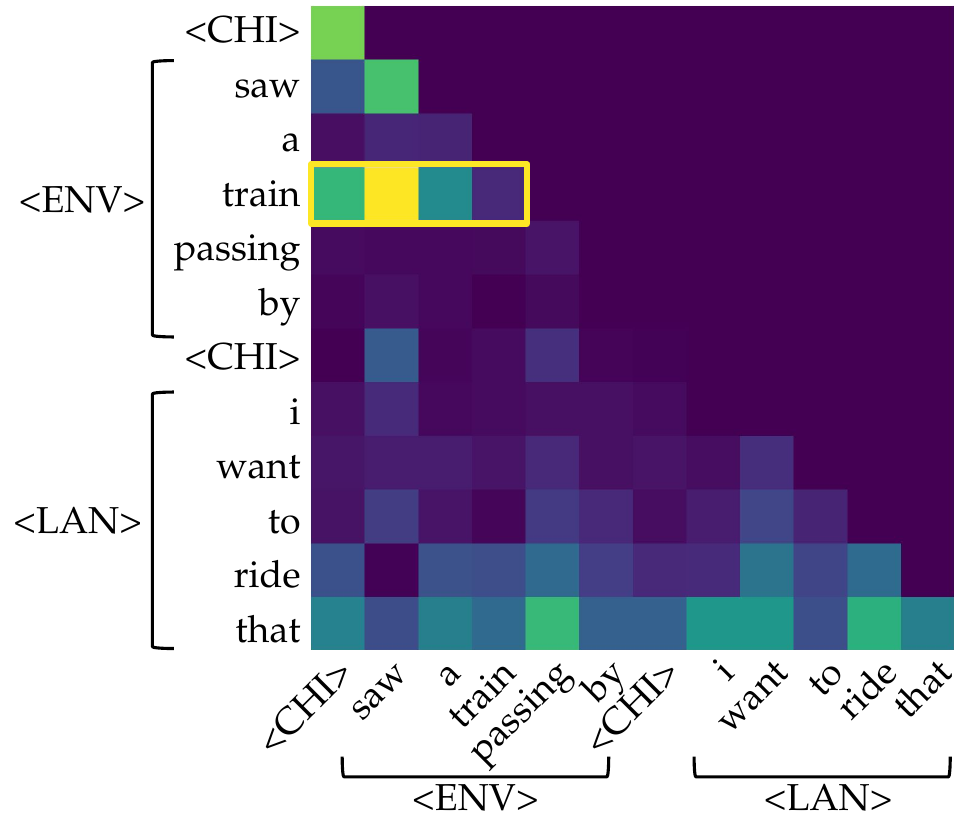}
        \vspace*{-10pt}
        \caption{Gather: L4 H7.}
        \label{fig:gather-1}
    \end{subfigure}
    \hfill
    \begin{subfigure}[t]{.48\linewidth}
        \centering
        \includegraphics[width=\linewidth]{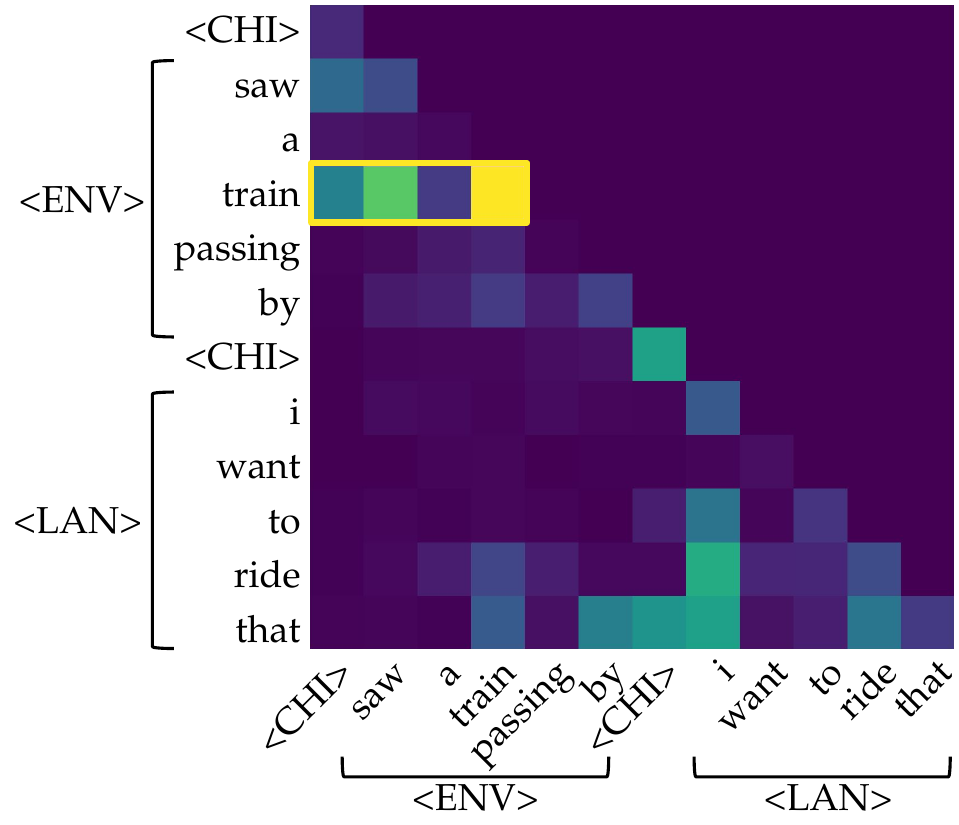}
        \vspace*{-10pt}
        \caption{Gather: L4 H8.}
        \label{fig:gather-2}
    \end{subfigure}

    \vspace{2pt}

    \begin{subfigure}[t]{.48\linewidth}
        \centering
        \includegraphics[width=\linewidth]{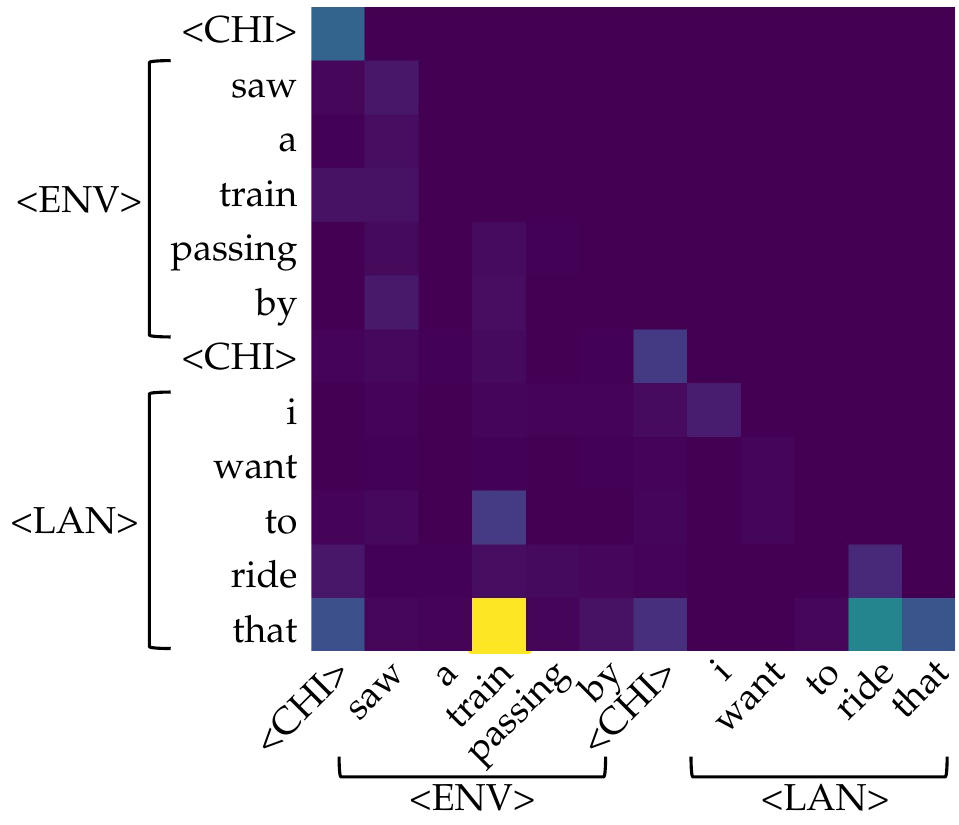}
        \vspace*{-10pt}
        \caption{Aggregate: L7 H5.}
        \label{fig:aggregate-1}
    \end{subfigure}
    \hfill
    \begin{subfigure}[t]{.48\linewidth}
        \centering
        \includegraphics[width=\linewidth]{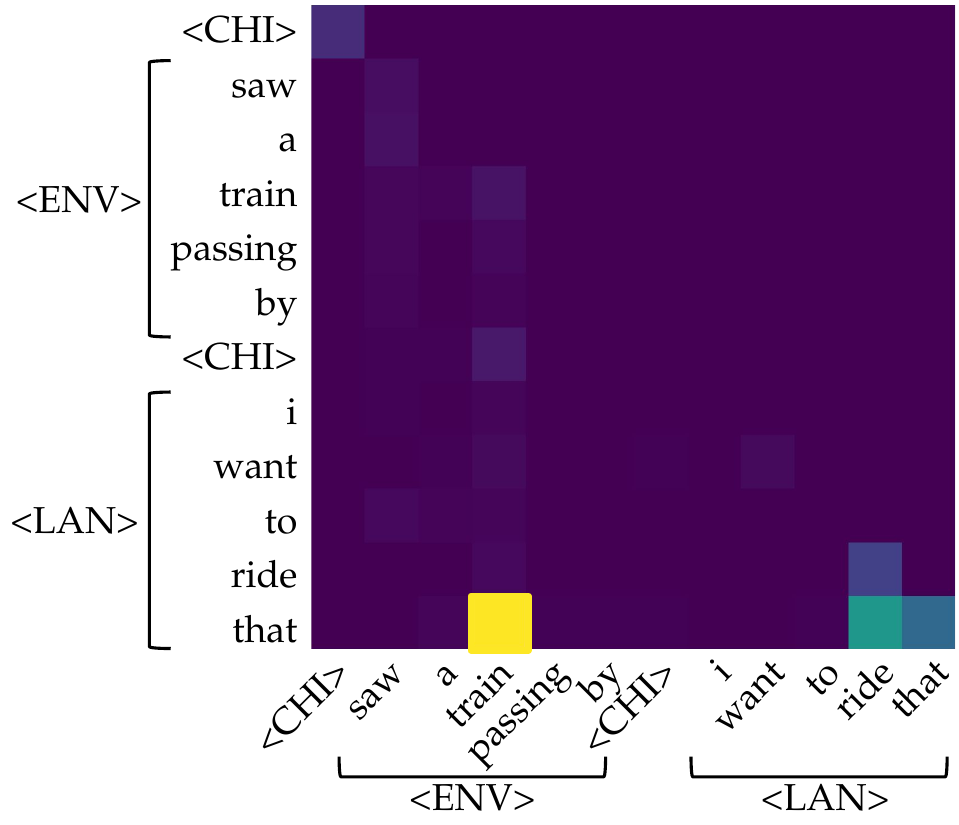}
        \vspace*{-10pt}
        \caption{Aggregate: L8 H5.}
        \label{fig:aggregate-2}
    \end{subfigure}

    \vspace*{-5pt}
    \caption{Examples of gather and aggregate heads identified. L: layer; H: head.}
    \label{fig:gna}
\end{figure}

Building on these results, we hypothesize that specific Transformer heads in the middle layers enable symbol grounding. 
To test the hypothesis, we examine attention saliencies for selected heads (Figure~\ref{fig:gna}).
We find that several heads exhibit patterns consistent with the gather and aggregate mechanisms described by \citet{bick2025understanding}: gather heads (e.g., Figures~\ref{fig:gather-1} and \ref{fig:gather-2}) compress relevant information into a subset of positions, while aggregate heads (e.g., Figures~\ref{fig:aggregate-1} and \ref{fig:aggregate-2}) redistribute this information to downstream tokens.
In our setups, saliency often concentrates on environmental tokens such as \textit{train}$_\env$, where gather heads pool contextual information into compact, retrievable states. 
In turn, aggregate heads broadcast this information from environmental ground (\textit{train}$\env$) to the token immediately preceding the linguistic form, thereby supporting the prediction of \textit{train}$_\lan$.
Taking these observations together, we hypothesize that the gather-and-aggregate heads implement the symbol grounding mechanism.

\begin{table}[!t]
\centering
\caption{Causal intervention results on identified gather and aggregate heads across training checkpoints (ckpt.), with a threshold value of 30\% as defined in section~\ref{sec:causal-intervention}. \textbf{Avg. Count} denotes the average number of heads of each type over inference times, and \textbf{Avg. Layer} denotes the average layer index where they appear. \textbf{Interv. Sps.} reports surprisal after zeroing out the identified heads, while \textbf{Ctrl. Sps.} reports surprisal after zeroing out an equal number of randomly selected heads. 
\textbf{Original} refers to the baseline surprisal without any intervention. 
{\scriptsize{***}} indicates a significant result ($p<0.001$) where the intervention surprisal is higher than that in the corresponding control experiment.
}
\label{tab:causal}
\vspace{-1pt}
\centering
\scalebox{.8}{
    \begingroup
    \setlength{\tabcolsep}{1.7pt}
    \renewcommand{\arraystretch}{0.75}
    \hspace*{-1pt}
    \begin{tabular}{cccccccccc}
    \toprule
    \multirow{3}{*}{Ckpt.} & \multicolumn{4}{c}{Gather Head} & \multicolumn{4}{c}{Aggregate Head} & \multirow{3}{*}{Original} \\
    \cmidrule(lr){2-5}\cmidrule(lr){6-9}
          & Avg. & Avg. & Interv. & Ctrl. & Avg. & Avg. & Interv. & Ctrl. &      \\
          & Count & Layer & Sps. & Sps. & Count & Layer & Sps. & Sps. &      \\
    \cmidrule(lr){1-1}\cmidrule(lr){2-3}\cmidrule(lr){4-5}\cmidrule(lr){6-7}\cmidrule(lr){8-9}\cmidrule(lr){10-10}
    500   & 0.00        & -   & -          & -       & 0.07  & 8.74              & 9.34    & 9.34 & 9.34 \\[2pt]
    5k  & 0.35     & 3.32  & 6.37         & 6.38    & 2.28     & 7.38  & \textbf{6.51 \scriptsize{(***)}}     & 6.39    & 6.38 \\[3pt]
    10k & 3.26     & 3.67  & 5.25         & 5.32    & 5.09     & 7.28  & \textbf{5.86 \scriptsize{(***)}}     & 5.29    & 5.30 \\[3pt]     
    20k & 5.76     & 3.59  & 4.69         & 4.79    & 6.71     & 7.52  & \textbf{5.62 \scriptsize{(***)}} & 4.76    & 4.77 \\ 
    \bottomrule
    \end{tabular}
\endgroup}
\vspace{-15pt}
\end{table}

\subsection{Causal Interventions of Attention Heads}
\label{sec:causal-intervention}

We then conduct causal interventions of attention heads to validate our previous hypothesis.

\boldstart{Operational definition.} 
We identify attention heads as gather or aggregate following these standards:
\begin{itemize}[topsep=-2pt,itemsep=0pt]
    \item \textbf{Gather head.} 
    An attention head is classified as a gather head if at least 30\% of its total saliency is directed toward the environmental ground token from the previous ones.
    \item \textbf{Aggregate head}: 
    An attention head is classified as an aggregate head if at least 30\% of its total saliency flows from the environmental ground token to the token immediately preceding the corresponding linguistic token.
\end{itemize}

\boldstart{Causal intervention methods.}
In each context, we apply causal interventions to the identified head types and their corresponding controls. 
Following \citet{bick2025understanding}, interventions are implemented by zeroing out the outputs of heads.
For the control, we mask an equal number of randomly selected heads in each layer, ensuring they do not overlap with the identified gather or aggregate heads.

\boldstart{Results and discussions.} 
As training progresses, the number of both gather and aggregate heads increases (Table~\ref{tab:causal}), suggesting that these mechanisms emerge over the course of learning.
Causal interventions reveal a clear dissociation: zeroing out aggregate heads consistently produces significantly higher surprisal compared to controls, whereas the gather head interventions have no such effect. 
This asymmetry suggests that gather heads serve in a role less critical in our settings, where the input template is semantically light and the environmental evidence alone suffices to shape the linguistic form. 
Layer-wise patterns further support this division of labor: gather heads cluster in shallow layers (3-4), while aggregate heads concentrate in mid layers (7-8). 
This resonates with our earlier probing results, where surprisal reductions became prominent only from layers 7-9. 
Together, these findings highlight aggregate heads in the middle layers as the primary account of grounding in the model.

% \vspace{-5pt}
% \input{floating/fig-gather-and-aggregate-vlm-tab-only}

\subsection{Generalization to Visual Dialog with Images} \label{sec:mechanistic-vlm}
% \vspace{-5pt}

We also conduct causal interventions on the VLM's attention heads to further validate our hypothesis above.

\boldstart{Operational definition.}
We define an attention head as an aggregate head if at least a certain portion (70\% or 90\% in our experiments) of its total image-to-text saliency flows from patches within the bounding box to the token immediately preceding the corresponding linguistic token.

% We identify an aggregate head if a certain portion (70\% or 90\% in our experiment settings) of its total image patch to end saliency flows from the patches inside bounding box to the token immediately preceding the corresponding linguistic token.

\boldstart{Causal intervention methods.}
Similarly to Section~\ref{sec:causal-intervention}, we apply causal interventions to the identified aggregate heads and their corresponding controls, by zeroing out their outputs.
For the control, we mask an equal number of randomly selected heads in each layer, ensuring they do not overlap with the identified aggregate heads.
\begin{figure}[!t]
\centering 
\begin{minipage}{0.5\textwidth}
\vspace{-1pt}
\scalebox{1}{
    \begingroup
    \setlength{\tabcolsep}{4pt}
    \renewcommand{\arraystretch}{0.7}
    \hspace*{-10pt}
    \begin{tabular}{ccccccc}
    \toprule
    \multirow{3}{*}{Thres.} & \multirow{3}{*}{Ckpt.} & \multicolumn{4}{c}{Aggregate Head} & \multirow{3}{*}{Original} \\
    \cmidrule(lr){3-6}
          &     & Avg. & Avg. & Interv. & Ctrl. &      \\
          &     & Count & Layer & Sps. & Sps. &      \\
    \cmidrule(lr){1-1}\cmidrule(lr){2-2}\cmidrule(lr){3-6}\cmidrule(lr){7-7}
    \multirow{7}{*}{70\%} & 20k  & 32.30    & 7.78  & 9.96         & 9.95    & 9.21 \\ [2pt]
                         & 100k & 35.63    & 7.71  & \textbf{9.42}         & 8.84    & 8.24 \\ [-2.5pt]
                         &      &          &       &
                         \scriptsize (***)         &     
                           &      \\
                         & 200k & 34.99    & 7.80   & \textbf{8.95}         & 8.15    & 7.76 \\ [-2.5pt]
                             &      &          &       &
                         \scriptsize (***)         &     
                           &      \\
                         & 300k & 34.15    & 7.76  & \textbf{8.96}         & 8.11    & 7.69 \\ [-2.5pt]
                           &      &          &       &
                         \scriptsize (***)         &     
                           &      \\
    \cmidrule(lr){1-1}\cmidrule(lr){2-2}\cmidrule(lr){3-6}\cmidrule(lr){7-7}
    \multirow{7}{*}{90\%} & 20k  & 10.66    & 8.33  & \textbf{9.51}         & 9.43    & 9.21 \\ [-2.5pt]
    &     &           &       &   \scriptsize (***)  &          &      \\
                         & 100k & 13.90     & 8.26  & \textbf{8.95}         & 8.50     & 8.24 \\ [-2.5pt]
                         &     &           &       &   \scriptsize (***)  &          &      \\
                         & 200k & 13.47    & 8.46  & \textbf{8.41}         & 7.88    & 7.76 \\ [-2.5pt]
                         &     &           &       &   \scriptsize (***)  &          &      \\
                         & 300k & 12.73    & 8.42  & \textbf{8.40}          & 7.87    & 7.69 \\ [-2.5pt]
                         &     &           &       &   \scriptsize (***)  &          &      \\
    \bottomrule
\end{tabular}
\endgroup}
\end{minipage}
\begin{minipage}{0.48\textwidth}
\centering
    \includegraphics[width=1.04\columnwidth]{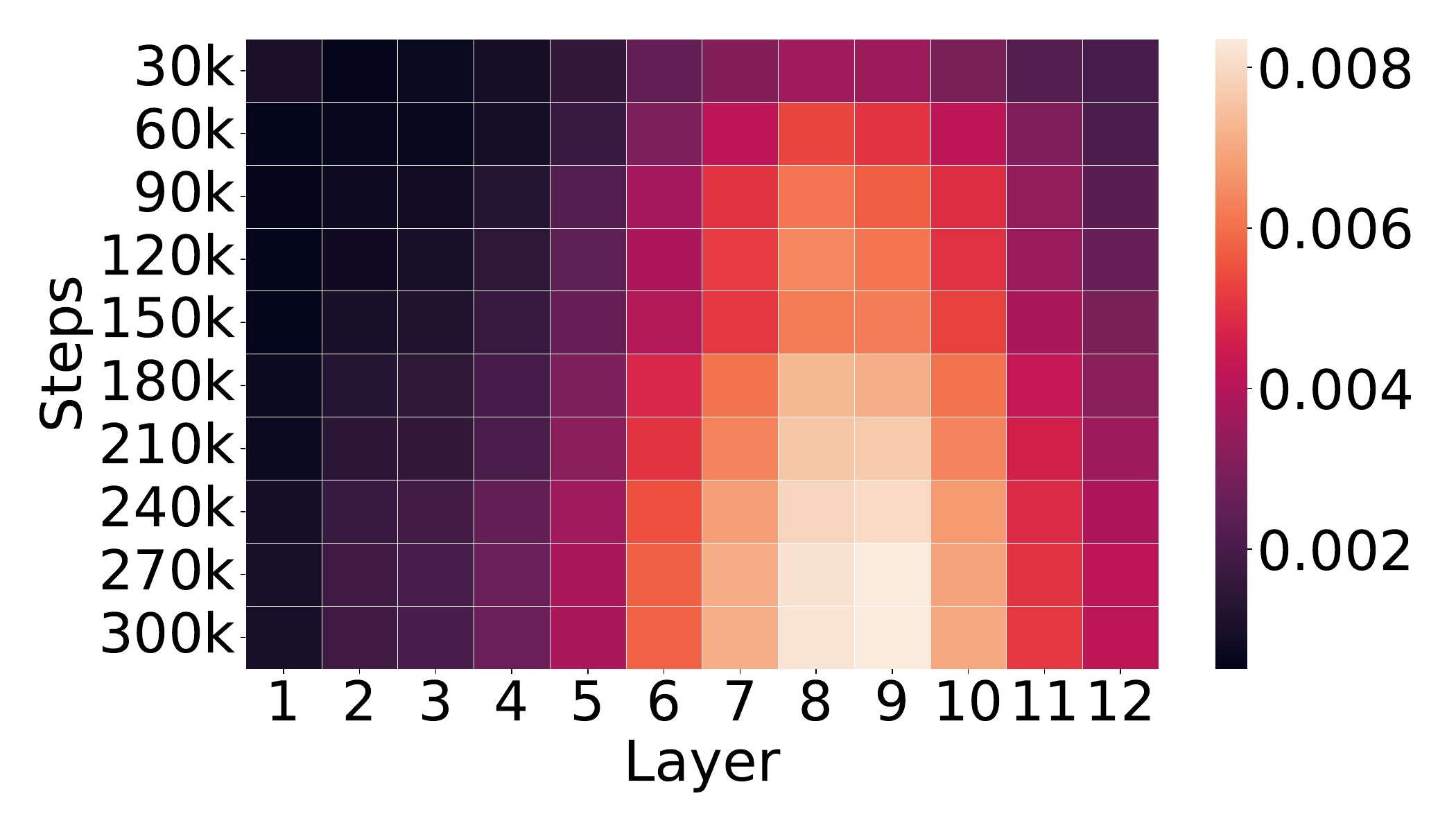}
\label{fig:vlm-saliency}
\end{minipage}
\vspace{-20pt}
\caption{Mechanistic analysis in the image-grounded visual dialogue setting. Top: Causal intervention results on identified aggregate heads across training checkpoints, where intervention on aggregate heads consistently yields significantly higher surprisal ($p<0.001$, ***) compared to the control group ones. 
Bottom: Saliency of layer-wise attention from environmental tokens (i.e., image tokens corresponding to patches within the bounding boxes of the target object) to linguistic tokens across training steps.
}
\label{fig:vlm-interp}
\end{figure}

\boldstart{Results and discussions.} 
As training progresses, the number of aggregate heads increases first and then becomes steady (Figure~\ref{fig:vlm-interp}, top), suggesting that these mechanisms emerge over the course of learning.
Zeroing out aggregate heads consistently produces significantly higher surprisal compared to the controls. 
The average layer depth of these heads also aligns well with the saliency heatmap (Figure~\ref{fig:vlm-interp}, bottom).

\section{Discussions}
\label{sec:discuss}

% \jycc{This is the only place the teaser figure Figure 1 is mentioned. I think we remove that figure from the second page. It can be distracting without the explanation. Part of it can move here to show an example of LLaVA. It will then make sense after reading through the paper up to this point. }
\hspace{-2pt}~\boldstart{Generalization to full-scale VLMs.}
As an additional case study, we extend our grounding-as-aggregation hypothesis to a full-scale VLM, LLaVA-1.5-7B \citep{liu2023visual}. 
Even in this heavily engineered architecture, we identify many attention heads exhibiting aggregation behavior consistent with our earlier findings (Figure~\ref{fig:overview_w_v}), reinforcing the view that symbol grounding arises from specialized heads.
% At the same time, full-scale VLMs present additional complications. 
% \jycc{this is not just a case study, this is a systematic study using the ChILD and language models. ``case study'' gives people the impression that it is only based on some cherry-picking examples. }
Meanwhile, full-scale models like LLaVA introduce additional complications: they incorporate CLIP-derived embeddings that already encode language priors for better performance, and global information may be stored in tokens out of the object regions~\citep{darcet2024vision}. 
Moreover, the large number of visual tokens substantially increases both computational cost and the difficulty of isolating genuine aggregation heads. 
These factors make systematic identification and intervention at scale a nontrivial challenge.
For these reasons, while our case study highlights promising evidence of grounding heads in modern VLMs, systematic detection and causal evaluation of such heads at scale remains an open challenge. 
Future work will need to develop computationally viable methods for detecting aggregation heads and applying causal interventions to validate their roles. 
% Addressing these challenges will be crucial for moving from anecdotal case studies to a more principled understanding of grounding in modern VLMs.

\boldstart{Connection to philosophical conceptualizations.}
While the primary purpose of our minimal testbed (child-direct speech setting; Table~\ref{tab:data}) is to offer a proxy for understanding grounding, experiments in this setting can be naturally viewed as investigations of symbol binding in language models, the problem that studies how symbols are connected together. 
This work extends the activation-based study on symbol binding \citep{feng2024how,dai-etal-2024-representational,feng2025monitoring} and offers evidence on the attention-head level, showing that aggregate heads are crucial in implementing the mechanism (Table~\ref{tab:causal}).
In line with \citet{yang2025emergent}, our work suggests that attention heads are crucial for implementing symbolic structures in LMs, and provides more controlled and causal evidence. 
% We offer a more in-depth discussion on the implications of this work to the philosophical roots of symbol grounding, in  Appendix~\ref{sec:phil}.

% Recent mechanistic studies have explored how LLMs bind entities and attributes in context. \citep{feng2024how, yang2025emergent}
\iffalse 
Recent mechanistic studies suggest that LLMs bind entities and attributes in context through specific internal representations, namely \textit{Binding IDs}, to track entity-attribute associations within the activation space~\citep{feng2024how}.
This understanding is further refined by evidence that LLMs also encode \textit{Ordering IDs} to track input sequences within the activation space~\citep{dai-etal-2024-representational}.
The existence of such bound states is further validated by propositional probes designed to monitor latent world states within the model's activations~\citep{feng2025monitoring}. 
Building directly on the binding ID mechanism, other work proposes an emergent symbolic architecture that supports reasoning in LLMs, hypothesizing that abstract variables are bound to tokens via similar mechanisms to support reasoning~\citep{yang2025emergent}.
\fi 

\boldstart{The Philosophical Roots of Grounding.}
Our findings highlight the need to sharpen the meaning of grounding in multimodal models. 
Prior work has often equated grounding with statistical correlations between visual and textual signals, such as attention overlaps or geometric alignments~\citep{cao2024emerging,bousselham2024grounding,schnaus2025s}.
While informative, such correlations diverge from the classic formulation by \citet{harnad1990symbol}, which requires symbols to be causally anchored to their referents in the environment.
In line with \citet{harnad1990symbol}, we frame grounding as a mechanistic property: one that can be traced along training, observed in the specialization of attention heads, and validated through causal interventions, providing a protocol for diagnosing when and how models genuinely tie symbols to meaning rather than mere correlations.
On another line, our results, which show that aggregate heads implement symbol grounding (and binding), echo \citet{pavlick2023symbolandgrounding} in arguing that \textit{LLMs lack the capacity to represent abstract symbolic structure} should not be accepted a priori. 
Instead, such claims should be evaluated carefully and empirically, with the focus on uncovering the models’ underlying competence rather than drawing conclusions solely from high-level architectures and surface-level performance.

% Our work provides a protocol for diagnosing when and how models genuinely tie symbols to meaning rather than merely exploiting correlations.
% From this perspective, grounding is not simply the co-occurrence of modalities, but the presence of identifiable mechanisms that link perceptual evidence to linguistic outputs in a way that is both necessary and predictive for correct generation. 
% This interpretation not only aligns grounding with its original theoretical roots, but also provides a framework for diagnosing when and how models genuinely tie symbols to meaning rather than merely exploiting correlations.

\boldstart{Practical implications to LM hallucinations.}
Our findings have practical implications for improving the reliability of LM outputs: by identifying aggregation heads that mediate grounding between environmental and linguistic tokens, we provide a promising mechanism to detect model reliability before generation.
Our findings echo a pathway to mitigate hallucinations by focusing on attention control: many hallucination errors stem from misallocated attention in intermediate layers~\citep{jiang2025devils,chen2024multi}.
Such attention-level signals can serve as early indicators of overtrust or false grounding, motivating practical solutions like decoding-time strategies to mitigate and eventually prevent hallucination~\citep{huang2024opera}.

\section*{Acknowledgements}
This work was supported in part by NSF IIS-1949634, NSF SES-2128623, NSERC RGPIN-2024-04395, the Weinberg Cognitive Science Fellowship to Ziqiao Ma, a Vector Scholarship to Xiaoxi Luo, and a Canada CIFAR AI Chair award to Freda Shi.
The authors would like to thank Songlin Yang and Jing Ding for their valuable feedback.

\section*{Impact Statement}
This study analyzes the mechanistic emergence of symbol grounding in (multimodal) language models using publicly available datasets such as CHILDES and Visual Dialog, and offers an example of pipelining interpretability techniques to understand large-scale neural networks. All the images are from the publicly available MSCOCO dataset \citep{lin2014microsoft}, where no personally identifiable data is used.

\iffalse
\section*{Reproducibility Statement}
We describe dataset construction, tokenization schemes, evaluation protocols, and model training settings in detail within the main text and Appendix. All experiments were repeated with multiple random seeds, and results across different architectures are reported. Code and data processing scripts are included in supplementary materials and will be released upon acceptance to facilitate full reproducibility.
\fi


\newpage
\bibliography{references}
\bibliographystyle{icml2026}

%%%%%%%%%%%%%%%%%%%%%%%%%%%%%%%%%%%%%%%%%%%%%%%%%%%%%%%%%%%%%%%%%%%%%%%%%%%%%%%
%%%%%%%%%%%%%%%%%%%%%%%%%%%%%%%%%%%%%%%%%%%%%%%%%%%%%%%%%%%%%%%%%%%%%%%%%%%%%%%
% APPENDIX
%%%%%%%%%%%%%%%%%%%%%%%%%%%%%%%%%%%%%%%%%%%%%%%%%%%%%%%%%%%%%%%%%%%%%%%%%%%%%%%
%%%%%%%%%%%%%%%%%%%%%%%%%%%%%%%%%%%%%%%%%%%%%%%%%%%%%%%%%%%%%%%%%%%%%%%%%%%%%%%
\newpage
\appendix
\onecolumn

\section{Dataset Details}
\label{app:data}

\subsection{Context Templates (Inference Time)}
We select the target tokens following the given procedure:
\begin{enumerate}
    \item Get a list of words, with their ENV and LAN frequency both greater than or equal to 100 in the CHILDES dataset;
    \item Get another list of nouns from CDI;
    \item Take intersection and select top 100 words (by frequency of their ENV token) as target token list.
\end{enumerate}

In CHILDES, all contexts are created with \texttt{gpt-4o-mini} followed by human verification if the genrated contexts are semantically light. Notice that these contexts are only used during inference.
We adopt the following prompt:

\definecolor{UMichBlue}{HTML}{00274C}
% \begin{verbatim}

\begin{figure}[!h]
    \centering
    \begin{tcolorbox}[colback=blue!5!white,colframe=UMichBlue,title=Prompt Templates for CHILDES]
    \begin{verbatim}
Given the word “{word}”, create 3 pairs of sentences that follow this 
requirement: 
1. The first sentence has a subject “The child”, describing an event or 
situation, and has the word “{word}”. Make sure to add a newline to the end of 
this first sentence
2. The second sentence is said by the child (only include the speech itself, 
don't include “the child say”, etc.), and the word “{word}” also appears in 
the sentence said by the child. Do not add quote marks either
3. Print each sentence on one line. Do not include anything else. 
4. Each sentence should be short, less than 10 words.
5. The word “{word}” in both sentence have the same meaning and have a clear 
indication or an implication relationship.
6. “{word}” should not appear at the first/second word of each sentence.
Generate 3 pairs of such sentences, so there should be 6 lines in total. 
You should not add a number. 
For each line, just print out the sentence.
\end{verbatim}
\end{tcolorbox}
    
\end{figure}

% In visual dialogue (caption version), we pre-define 10 sets of templates directly:
In visual dialogue (caption version and VLM version), we pre-define 10 sets of templates for each version:
\begin{figure}[!h]
    \centering
    \begin{tcolorbox}[colback=blue!5!white,colframe=UMichBlue,title=Prompt Templates for Visual Dialogue (Caption Version)]
\begin{verbatim}
this:<ENV> is:<ENV> [FILLER]:<ENV> <Q> what:<LAN> is:<LAN> it:<LAN> <A> 
(predict [FILLER]:<LAN>)

this:<ENV> is:<ENV> [FILLER]:<ENV> <Q> what:<LAN> do:<LAN> you:<LAN> 
call:<LAN> this:<LAN> <A> (predict [FILLER]:<LAN>)

this:<ENV> is:<ENV> [FILLER]:<ENV> <Q> can:<LAN> you:<LAN>
name:<LAN> this:<LAN> object:<LAN> <A> 
(predict [FILLER]:<LAN>)

this:<ENV> is:<ENV> [FILLER]:<ENV> <Q> what's:<LAN> 
this:<LAN> called:<LAN> <A> 
(predict [FILLER]:<LAN>)
\end{verbatim}

\end{tcolorbox}
    
\end{figure}

\begin{figure}[!h]
    \centering
    \begin{tcolorbox}[colback=blue!5!white,colframe=UMichBlue,title=Prompt Templates for Visual Dialogue (Caption Version) (continued)]
    \begin{verbatim}
this:<ENV> is:<ENV> [FILLER]:<ENV> <Q> what:<LAN> 
this:<LAN> thing:<LAN> is:<LAN> <A> 
(predict [FILLER]:<LAN>)

this:<ENV> is:<ENV> [FILLER]:<ENV> <Q> what:<LAN> 
would:<LAN> you:<LAN> name:<LAN> this:<LAN> <A> 
(predict [FILLER]:<LAN>)

this:<ENV> is:<ENV> [FILLER]:<ENV> <Q> 
what's:<LAN> the:<LAN> name:<LAN> of:<LAN> this:<LAN> 
item:<LAN> <A> (predict [FILLER]:<LAN>)

this:<ENV> is:<ENV> [FILLER]:<ENV> <Q> how:<LAN> 
do:<LAN> you:<LAN> identify:<LAN> this:<LAN> <A> 
(predict [FILLER]:<LAN>)

this:<ENV> is:<ENV> [FILLER]:<ENV> <Q> what:<LAN> 
do:<LAN> we:<LAN> have:<LAN> here:<LAN> <A> 
(predict [FILLER]:<LAN>)

this:<ENV> is:<ENV> [FILLER]:<ENV> <Q> how:<LAN> 
do:<LAN> you:<LAN> call:<LAN> this:<LAN> 
object:<LAN> <A> (predict [FILLER]:<LAN>)
    
\end{verbatim}
\end{tcolorbox}
    
\end{figure}

% In the VLM version, we pre-define 10 sets of templates:
\begin{figure}[!h]
    \centering
    \begin{tcolorbox}[colback=blue!5!white,colframe=UMichBlue,title=Prompt Templates for Visual Dialogue (VLM Version)]
\begin{verbatim}
“<image> \nwhat is it ?”,
“<image> \nwhat do you call this ?”,
“<image> \ncan you name this object ?”,
“<image> \nwhat is this called ?”,
“<image> \nwhat this thing is ?”,
“<image> \nwhat would you name this ?”,
“<image> \nwhat is the name of this item ?”,
“<image> \nhow do you identify this ?”,
“<image> \nwhat do we have here ?”,
“<image> \nhow do you call this object ?”
\end{verbatim}
\end{tcolorbox}
    
\end{figure}

% \subsection{Sample context}
% <CHI> played:<ENV> happily:<ENV> with:<ENV> a:<ENV> colorful:<ENV> toy:<ENV> <CHI> I:<LAN> love:<LAN> this:<LAN> (predict toy:<LAN>)

\subsection{Word List for CHILDES and Vision Dialogue (Text Only)}
\label{sec:word-transformer}
% [``box'', ``book'', ``ball'', ``hand'', ``paper'', ``table'', ``toy'', ``head'', ``car'', ``chair'', ``room'', ``picture'', ``doll'', ``cup'', ``towel'', ``door'', ``mouth'', ``camera'', ``duck'', ``face'', ``truck'', ``bottle'', ``puzzle'', ``bird'', ``tape'', ``finger'', ``bucket'', ``block'', ``stick'', ``elephant'', ``hat'', ``bed'', ``arm'', ``dog'', ``kitchen'', ``spoon'', ``hair'', ``blanket'', ``horse'', ``tray'', ``train'', ``cow'', ``foot'', ``couch'', ``necklace'', ``cookie'', ``plate'', ``telephone'', ``window'', ``brush'', ``ear'', ``pig'', ``purse'', ``hammer'', ``cat'', ``shoulder'', ``garage'', ``button'', ``monkey'', ``pencil'', ``shoe'', ``drawer'', ``leg'', ``bear'', ``milk'', ``egg'', ``bowl'', ``juice'', ``ladder'', ``basket'', ``coffee'', ``bus'', ``food'', ``apple'', ``bench'', ``sheep'', ``airplane'', ``comb'', ``bread'', ``eye'', ``animal'', ``knee'', ``shirt'', ``cracker'', ``glass'', ``light'', ``game'', ``cheese'', ``sofa'', ``giraffe'', ``turtle'', ``stove'', ``clock'', ``star'', ``refrigerator'', ``banana'', ``napkin'', ``bunny'', ``farm'', ``money'']
[box, book, ball, hand, paper, table, toy, head, car, chair, room, picture, doll, cup, towel, door, mouth, camera, duck, face, truck, bottle, puzzle, bird, tape, finger, bucket, block, stick, elephant, hat, bed, arm, dog, kitchen, spoon, hair, blanket, horse, tray, train, cow, foot, couch, necklace, cookie, plate, telephone, window, brush, ear, pig, purse, hammer, cat, shoulder, garage, button, monkey, pencil, shoe, drawer, leg, bear, milk, egg, bowl, juice, ladder, basket, coffee, bus, food, apple, bench, sheep, airplane, comb, bread, eye, animal, knee, shirt, cracker, glass, light, game, cheese, sofa, giraffe, turtle, stove, clock, star, refrigerator, banana, napkin, bunny, farm, money]

\subsection{Word List for Vision Dialogue (VLM)}
\label{sec:word-vlm}
% [``box'', ``book'', ``table'', ``toy'', ``car'', ``chair'', ``doll'', ``door'', ``camera'', ``duck'', ``truck'', ``bottle'', ``bird'', ``elephant'', ``hat'', ``bed'', ``dog'', ``spoon'', ``horse'', ``train'', ``couch'', ``necklace'', ``cookie'', ``plate'', ``telephone'', ``window'', ``pig'', ``cat'', ``monkey'', ``drawer'', ``bear'', ``milk'', ``egg'', ``bowl'', ``juice'', ``ladder'', ``bus'', ``food'', ``apple'', ``sheep'', ``bread'', ``animal'', ``shirt'', ``cheese'', ``giraffe'', ``clock'', ``refrigerator'', ``accordion'', ``aircraft'', ``alpaca'', ``ambulance'', ``ant'', ``antelope'', ``backpack'', ``bagel'', ``balloon'', ``barrel'', ``bathtub'', ``beard'', ``bee'', ``beer'', ``beetle'', ``bicycle'', ``bidet'', ``billboard'', ``boat'', ``bookcase'', ``boot'', ``boy'', ``broccoli'', ``building'', ``bull'', ``burrito'', ``bust'', ``butterfly'', ``cabbage'', ``cabinetry'', ``cake'', ``camel'', ``canary'', ``candle'', ``candy'', ``cannon'', ``canoe'', ``carrot'', ``cart'', ``castle'', ``caterpillar'', ``cattle'', ``cello'', ``cheetah'', ``chicken'', ``chopsticks'', ``closet'', ``clothing'', ``coat'', ``cocktail'', ``coffeemaker'', ``coin'', ``cosmetics'']
[box, book, table, toy, car, chair, doll, door, camera, duck, truck, bottle, bird, elephant, hat, bed, dog, spoon, horse, train, couch, necklace, cookie, plate, telephone, window, pig, cat, monkey, drawer, bear, milk, egg, bowl, juice, ladder, bus, food, apple, sheep, bread, animal, shirt, cheese, giraffe, clock, refrigerator, accordion, aircraft, alpaca, ambulance, ant, antelope, backpack, bagel, balloon, barrel, bathtub, beard, bee, beer, beetle, bicycle, bidet, billboard, boat, bookcase, boot, boy, broccoli, building, bull, burrito, bust, butterfly, cabbage, cabinetry, cake, camel, canary, candle, candy, cannon, canoe, carrot, cart, castle, caterpillar, cattle, cello, cheetah, chicken, chopsticks, closet, clothing, coat, cocktail, coffeemaker, coin, cosmetics]

\section{Implementation Details}

\subsection{Checkpointing}
\subsubsection{CHILDES, Visual Dialogue with captions}
We save the intermediate steps: [0, 150, 300, 500, 1000, 1500, 2000, 2500, 3000, 3500, 4000, 4500, 5000, 5500, 6000, 6500, 7000, 7500, 8000, 8500, 9000, 9500, 10000, 11000, 12000, 13000, 14000, 15000, 16000, 17000, 18000, 19000, 20000] (33 checkpoints in total)

\subsubsection{Visual Dialogue (VLM)}
We save the intermediate steps: [10000, 20000, 40000, 60000, 80000, 100000, 120000, 140000, 160000, 180000, 200000, 220000, 240000, 260000, 280000, 300000] (16 checkpoints in total)

% \subsection{Visual Dialogue Results.}
% We provide the Visual Dialogue results in Figure~\ref{fig:sps-visdial} and Figure~\ref{fig:r2-visdial}.

% \input{floating/behavior_visdial}
% \input{floating/cooccurrence_visdial}

\begin{table*}[!t]
    \caption{Training and test examples across datasets with target word \textcolor{blue}{\it book}. The training examples combine environmental tokens (\colorbox{tticblue!10}{\env; shaded}) with linguistic tokens (\lan). Test examples are constructed with either matched (\textcolor{blue}{\it book}) or mismatched (\textcolor{red}{\it toy}) environmental contexts, paired with corresponding linguistic prompts. 
    Note that in caption-grounded dialogue, \textcolor{blue}{\it book$_\env$} and \textcolor{blue}{\it book$_\lan$} are two distinct tokens received by LMs.
    }
    \vspace{-3pt}
    \label{tab:data-caption}
    \centering
    \scalebox{0.9}{
    \begingroup
    \setlength{\tabcolsep}{5.5pt}
    \hspace*{-7pt}
    \begin{tabular}{m{0.10\linewidth} m{0.17\linewidth} m{0.23\linewidth} m{0.15\linewidth} m{0.18\linewidth} m{0.11\linewidth}}
    \toprule
    \multirow{2}{*}{\parbox{20pt}{\bf Dataset}} & \multicolumn{2}{c}{\bf Training Example} & \multicolumn{3}{c}{\bf Test Example} \\
    \cmidrule(lr){2-3}\cmidrule(lr){4-6}
     &
    \multicolumn{1}{c}{\cellcolor{tticblue!10} \bf \env} &
    \multicolumn{1}{c}{\bf \lan} &
    \multicolumn{1}{c}{\cellcolor{tticblue!10} \bf \env Match} &
    \multicolumn{1}{c}{\cellcolor{tticblue!10} \bf \env Mismatch} &
    \multicolumn{1}{c}{\bf \lan} \\
    \midrule
    {\bf Caption-Grounded Dialogue} &
    \cellcolor{tticblue!10} \textit{a dog appears to be reading a \textcolor{blue}{book} with a full bookshelf behind} &
    \textit{$\langle$Q$\rangle$ can you tell what \textcolor{blue}{book} it's reading $\langle$A$\rangle$ the marriage of true minds by stephen evans} &
    \cellcolor{tticblue!10} \textit{this is a \textcolor{blue}{book}} &
    \cellcolor{tticblue!10} \textit{this is a \textcolor{red}{toy}} &
    \textit{$\langle$Q$\rangle$ can you name this object $\langle$A$\rangle$ \underline{\hspace{3em}}} \\
    \bottomrule
    \end{tabular}
    \endgroup}
    \vspace*{-10pt}
\end{table*}

\begin{figure*}[!t]
    \centering
    \begin{subfigure}[t]{.29\textwidth}
        \centering
        \includegraphics[width=1.03\columnwidth]{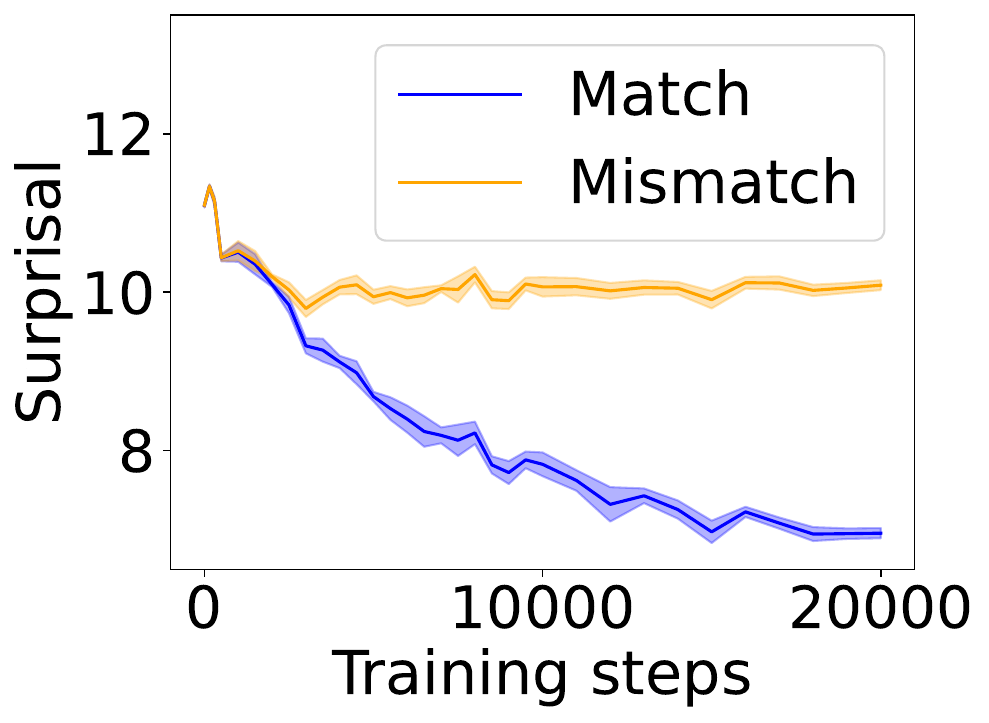}
        \vspace*{-15pt}
        \caption{Surprisal curves (w/ caption).}
        \label{fig:sps-visdial-cap}
    \end{subfigure}
    ~
    \begin{subfigure}[t]{.32\textwidth}
        \centering
        \includegraphics[width=1.04\columnwidth]{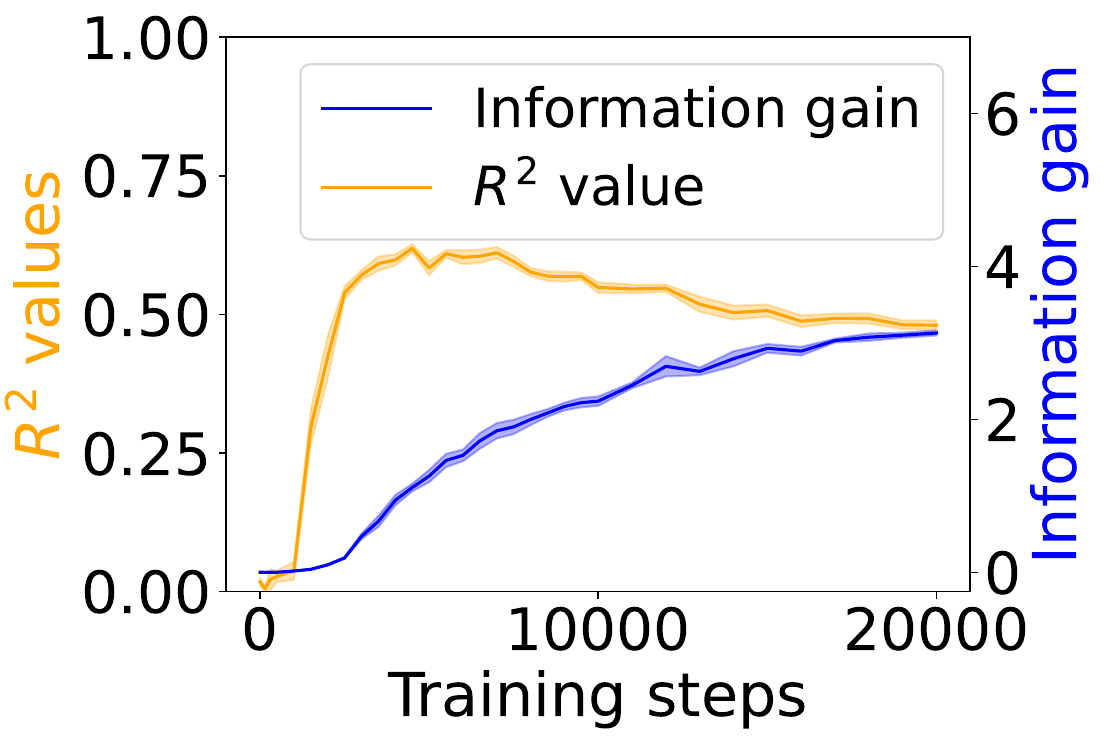}
        \vspace*{-15pt}
        \caption{$R^2$ and information gain (w/ caption).}
        \label{fig:r2-visdial-cap}
    \end{subfigure}
    \vspace*{-5pt}
    \caption{Average surprisal of the experimental and control conditions, as well as the grounding information gain and its correlation to the co-occurrence of linguistic and environment tokens over training steps. 
    All results are from a 12-layer Transformer model on grounded dialogue data. 
        \vspace*{-10pt}
    }
    \label{fig:visdial-caption}
\end{figure*}

\section{Addendum to Results}
\subsection{Caption-grounded dialogue}
\label{sec:Caption-grounded}
We use the same dataset as the image-grounded dialogue, the Visual Dialog dataset~\citep{das2017visual}. In our setup, MSCOCO captions serve as the environmental tokens (\env) and the dialogue turns form the linguistic tokens (\lan). 
In this pseudo cross-modal setting, textual descriptions of visual scenes ground natural conversational interaction. 
Compared to CHILDES, this setup introduces richer semantics and longer utterances, while still using text-based inputs for both token types, thereby offering a stepping stone toward grounding in fully visual contexts.

To assess symbol grounding in caption-grounded dialogue, we use a similar contrastive test as Child-directed speech and Image-grounded Dialogue, as demonstrated in Table \ref{tab:data}.

In this setting, we also train 12-layer Transformers with 5 random seeds.
Compared to Figures~\ref{fig:sps-visdial-img}--\ref{fig:r2-visdial-img}, there is a similar but stronger pattern: a larger surprisal gap exists between match and mismatch conditions (Figure~\ref{fig:sps-visdial-cap}), with the grounding information gain increasing steadily while $R^2$ peaks early and declines (Figure~\ref{fig:r2-visdial-cap}). 
Both settings confirm that emergent grounding cannot be fully explained by co-occurrence statistics.

\subsection{Detailed Behavioral Analysis for all Models}
\begin{figure*}[!t]
    \centering
    \begin{subfigure}[t]{0.32\textwidth}
        \centering
        \includegraphics[width=1.00\columnwidth]{figs/surprisal_overtime/childes_transformer_4layer.pdf}
        \vspace*{-15pt}
        \caption{4-layer Transformer.}
    \end{subfigure}
    ~
    \begin{subfigure}[t]{0.32\textwidth}
        \centering
        \includegraphics[width=1.00\columnwidth]{figs/surprisal_overtime/childes_transformer.pdf}
        \vspace*{-15pt}
        \caption{12-layer Transformer.}
    \end{subfigure}
    ~
        \begin{subfigure}[t]{0.32\textwidth}
        \centering
        \includegraphics[width=1.00\columnwidth]{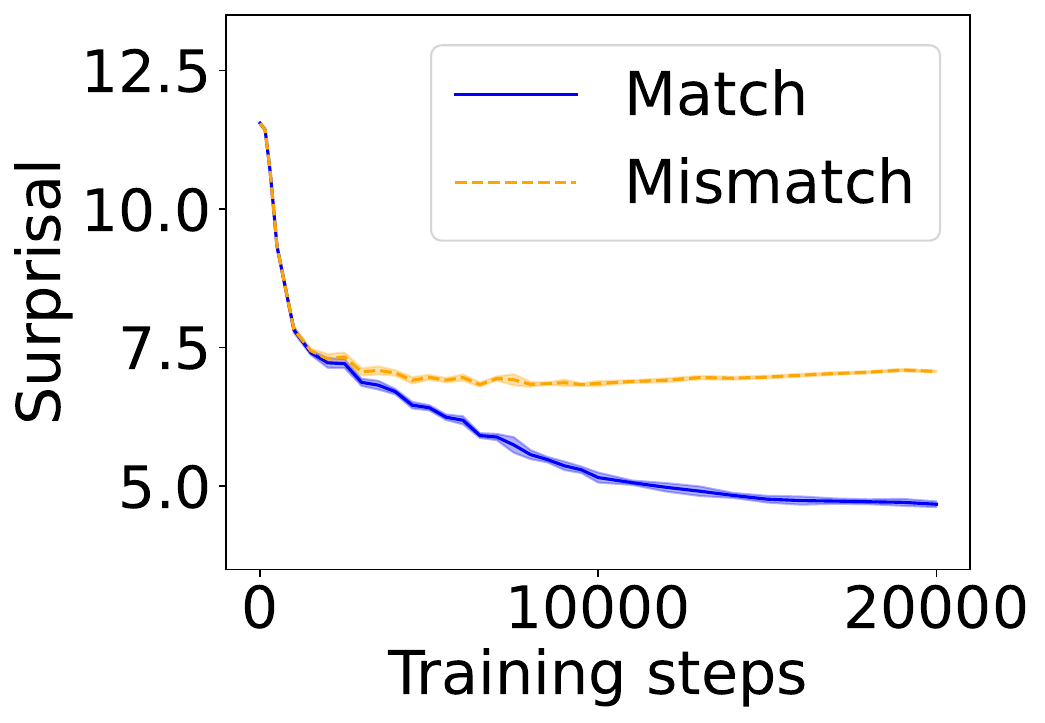}
        \vspace*{-15pt}
        \caption{18-layer Transformer.}
    \end{subfigure}
    ~
    \begin{subfigure}[t]{0.32\textwidth}
        \centering
        \includegraphics[width=1.00\columnwidth]{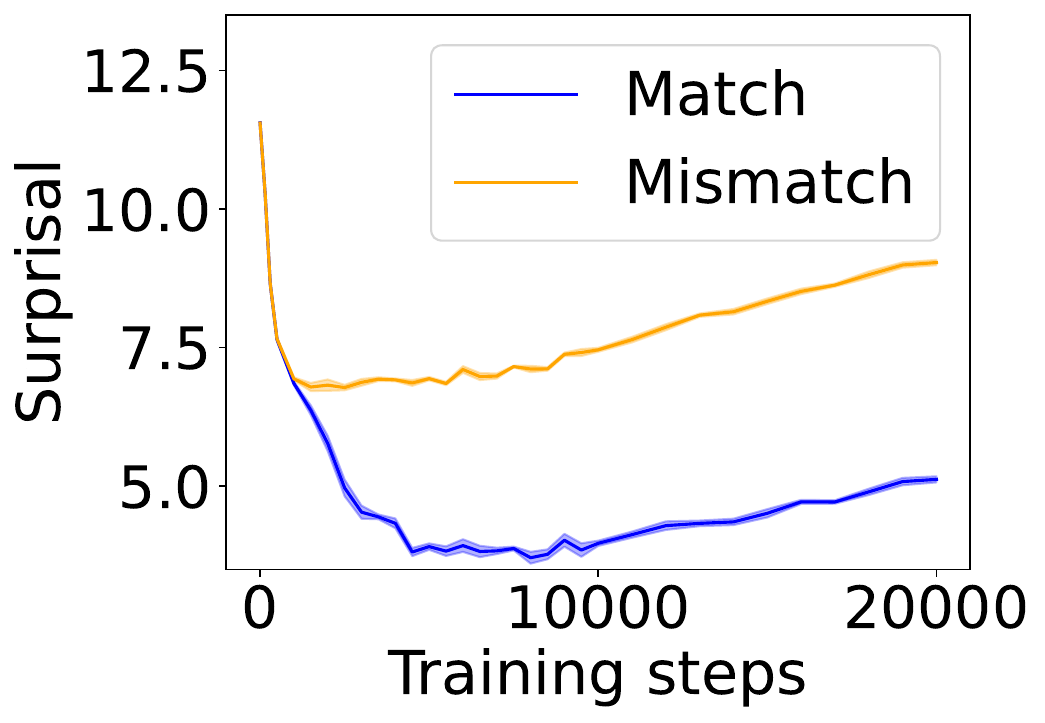}
        \vspace*{-15pt}
        \caption{12-layer Mamba 2.}
    \end{subfigure}
    ~
    \begin{subfigure}[t]{0.32\textwidth}
        \centering
        \includegraphics[width=1.00\columnwidth]{figs/surprisal_overtime/childes_mamba2_nores_4layer.pdf}
        \vspace*{-15pt}
        \caption{4-layer Mamba 2.}
    \end{subfigure}
    ~
    \begin{subfigure}[t]{0.32\textwidth}
        \centering
        \includegraphics[width=1.00\columnwidth]{figs/surprisal_overtime/childes_lstm.pdf}
        \vspace*{-15pt}
        \caption{4-layer LSTM.}
    \end{subfigure}
    \vspace*{-5pt}
    \caption{Average surprisal of the experimental and control conditions over training steps. \vspace*{-10pt}}
    \label{fig:surprisal-childes-all}
\end{figure*}

\begin{figure*}[!t]
    \centering
    \begin{subfigure}[t]{.32\textwidth}
        \centering
        \includegraphics[width=1.00\columnwidth]{figs/coocurrence/childes_transformer_4layer.pdf}
        \vspace*{-15pt}
        \caption{4-layer Transformer.}
    \end{subfigure}
    ~
    \begin{subfigure}[t]{.32\textwidth}
        \centering
        \includegraphics[width=1.00\columnwidth]{figs/coocurrence/childes_transformer.pdf}
        \vspace*{-15pt}
        \caption{12-layer Transformer.}
    \end{subfigure}
    ~
        \begin{subfigure}[t]{.32\textwidth}
        \centering
        \includegraphics[width=1.00\columnwidth]{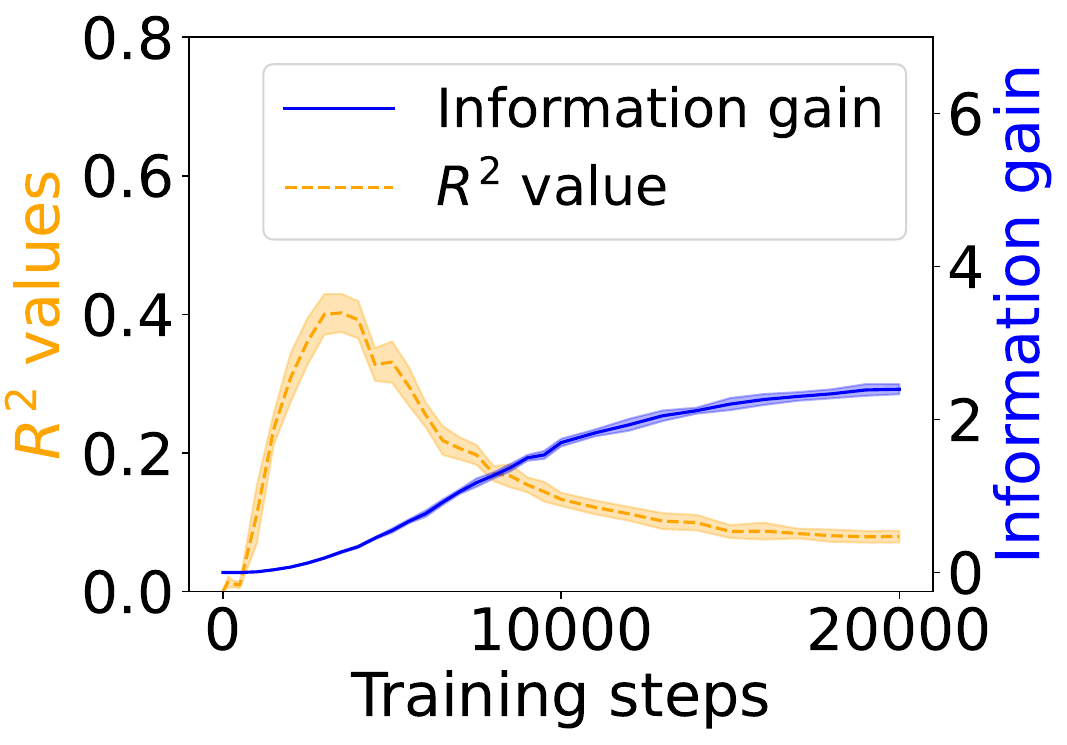}
        \vspace*{-15pt}
        \caption{18-layer Transformer.}
    \end{subfigure}
    ~
    \begin{subfigure}[t]{.32\textwidth}
        \centering
        \includegraphics[width=1.00\columnwidth]{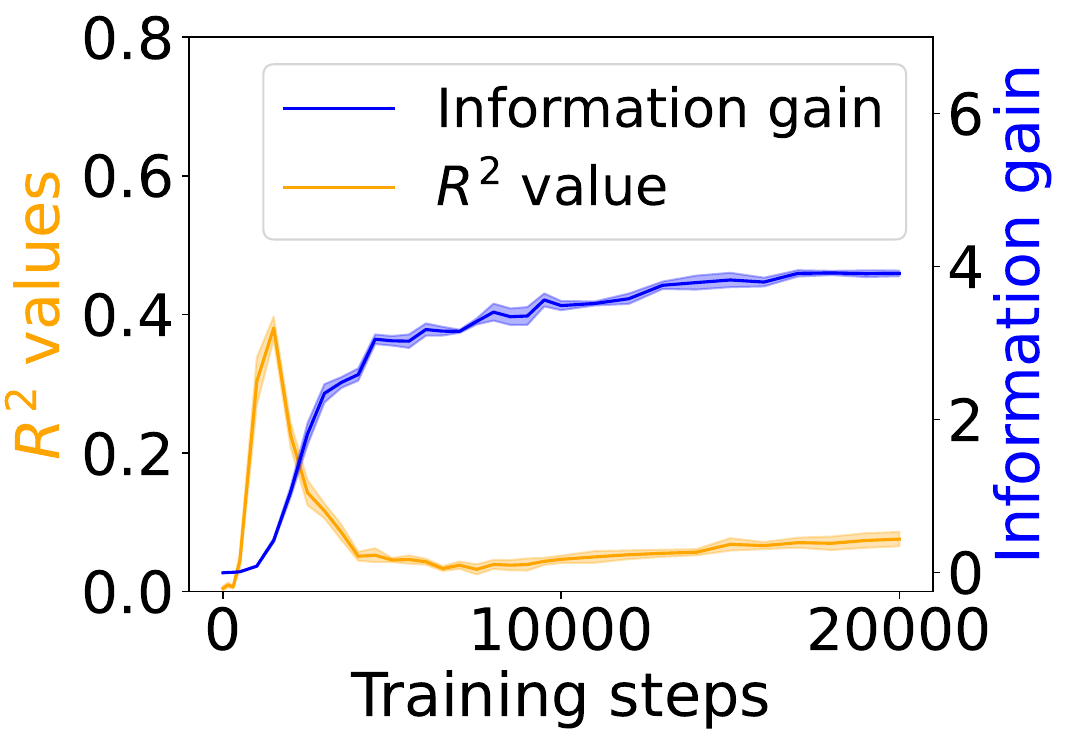}
        \vspace*{-15pt}
        \caption{12-layer Mamba 2.}
    \end{subfigure}
    ~
    \begin{subfigure}[t]{.32\textwidth}
        \centering
        \includegraphics[width=1.00\columnwidth]{figs/coocurrence/childes_mamba2_nores_4layer.pdf}
        \vspace*{-15pt}
        \caption{4-layer Mamba 2.}
    \end{subfigure}
    ~
    \begin{subfigure}[t]{.32\textwidth}
        \centering
        \includegraphics[width=1.00\columnwidth]{figs/coocurrence/childes_lstm.pdf}
        \vspace*{-15pt}
        \caption{4-layer LSTM.}
    \end{subfigure}
    \vspace*{-5pt}
    \caption{Grounding information gain and its correlation to the co-occurrence of linguistic and environment tokens over training steps. \vspace*{-10pt}}
    \label{fig:r2-childes-full}
\end{figure*}

\begin{figure*}[!t]
    \centering
    \begin{subfigure}[t]{0.77\textwidth}
        \centering
        \includegraphics[width=1.0\columnwidth]{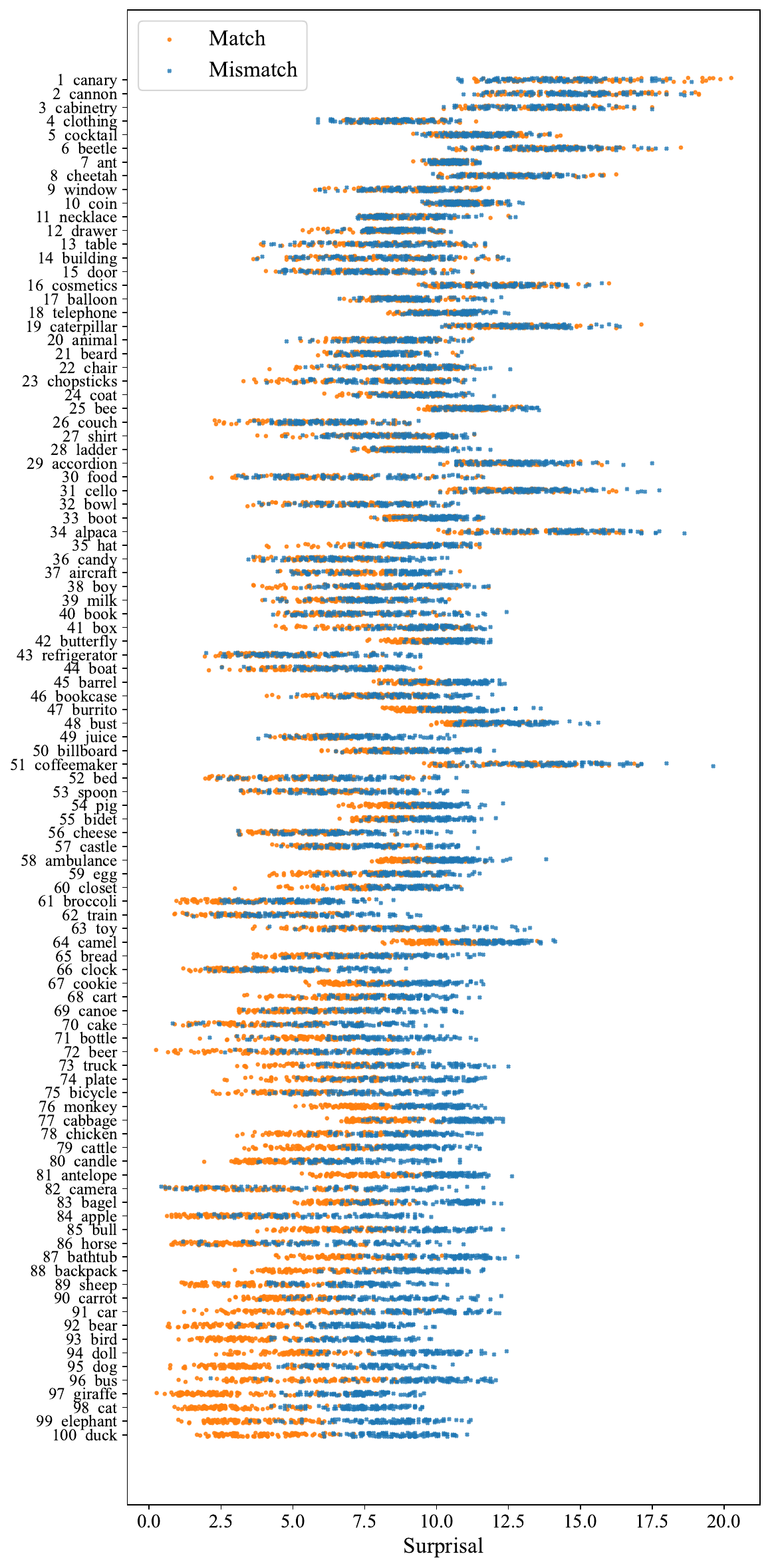}
    \end{subfigure}
    \vspace*{-15pt}
    \caption{
        Per-instance surprisal in our trained VLM, sorted by information-gain per word (increasing). 
        Orange dots: surprisals in matched context; blue cross: surprisals in mismatched context. 
    }
    \label{fig:vlm-beeswarm}
\end{figure*}

We show the complete behavioral evidence for all models in Figure~\ref{fig:surprisal-childes-all}, and co-occurrence analysis in Figure~\ref{fig:r2-childes-full}. 
On top of that, for 12-layer Transformers, a beeswarm plot indicating per context match/mismatch surprisal is shown in Figure~\ref{fig:vlm-beeswarm}.

\subsection{Detailed Gather and Aggregate Analysis (Transformer)}

\begin{figure*}[!t]
    \centering
    \begin{subfigure}[t]{.45\textwidth}
        \centering
        \includegraphics[width=1.03\columnwidth]{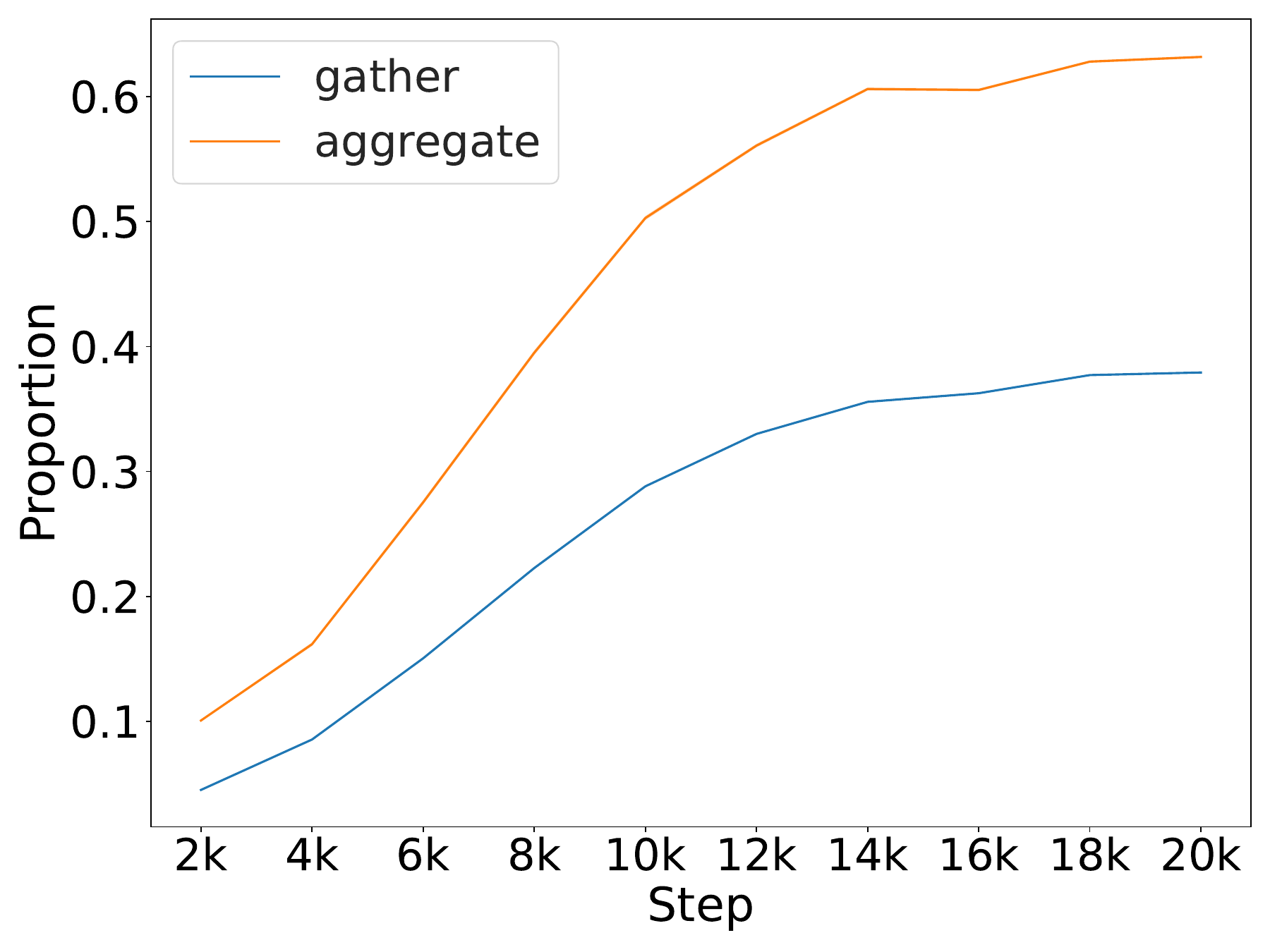}
    \end{subfigure}
    \vspace*{-8pt}
    \caption{Gather-and-aggregate overtime strength. \vspace*{-5pt}}
    \label{fig:ga-overtime}
\end{figure*}

After finding the set of gather and aggregate heads for each context, we run an overtime analysis showing the proportion of saliency to the total saliency, as is shown in Figure~\ref{fig:ga-overtime}.

\subsection{Relationship between Aggregate Head Number and Grounding Information Gain}
We examine the 12-layer Transformer LM for both the child-directed speech and VLM settings.
For each target token (detailed in Sections~\ref{sec:word-transformer} and \ref{sec:word-vlm}), we compare its grounding information gain with the average aggregate head it has (Figure~\ref{fig:ga-IG}). 
We find an intermediate-to-strong positive correlation between the grounding information gain and the aggregate head number detected. 
% The $R^2$ for Figure ~\ref{fig:ga-IG-childes} is 0.847, for Figure ~\ref{fig:ga-IG-vlm70} is 0.250, and for Figure ~\ref{fig:ga-IG-vlm90} is 0.175.

{
\color{blue}
\begin{figure*}[!t]
    \vspace*{-15pt}
    \centering
    \begin{subfigure}[t]{.265\textwidth}
        \centering
        \includegraphics[width=1.03\columnwidth]{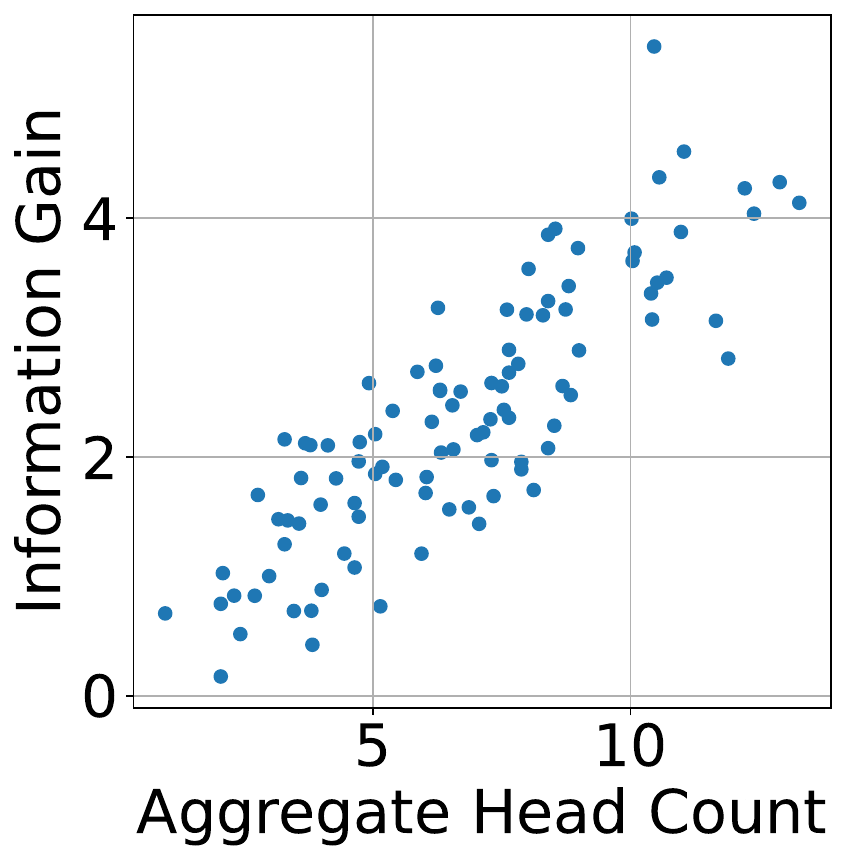}
        \vspace*{-15pt}
        \caption{Information Gain vs Aggregate Head Number (CHILDES), $R^2=0.847$.}
        \label{fig:ga-IG-childes}
    \end{subfigure}
    ~
    \begin{subfigure}[t]{.265\textwidth}
        \centering
        \includegraphics[width=1.03\columnwidth]{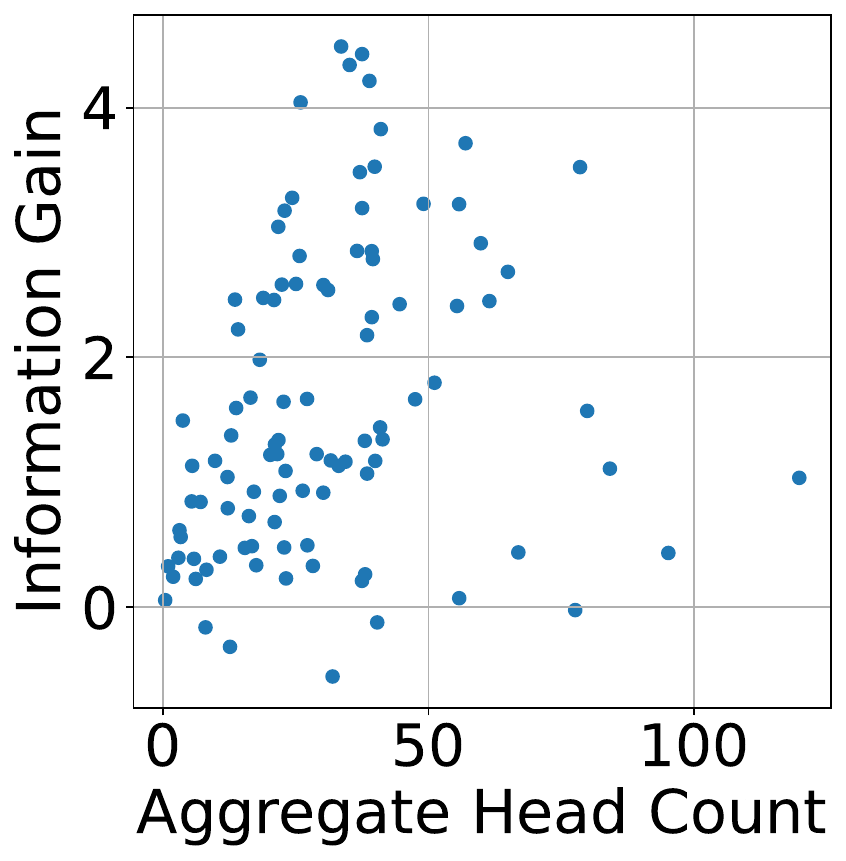}
        \vspace*{-15pt}
        \caption{Information Gain vs. Aggregate Head Number (VLM, with 70\% threshold), $R^2=0.250$.}
        \label{fig:ga-IG-vlm70}
    \end{subfigure}
    ~
    \begin{subfigure}[t]{.265\textwidth}
        \centering
        \includegraphics[width=1.03\columnwidth]{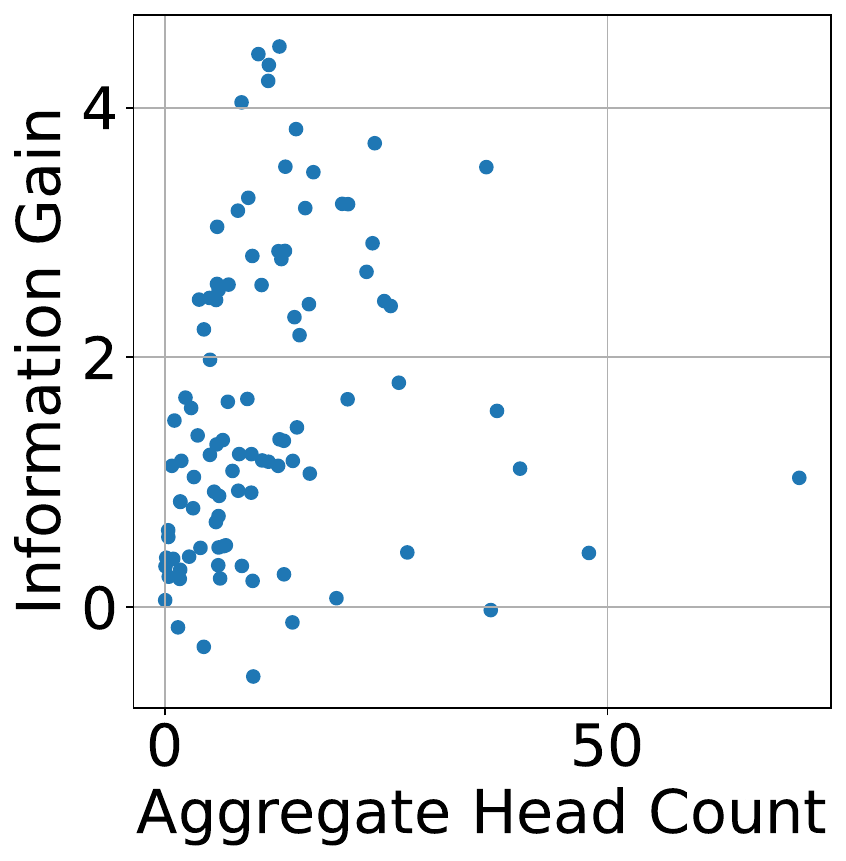}
        \vspace*{-15pt}
        \caption{Information Gain vs. Aggregate Head Number (VLM, with 90\% threshold), $R^2=0.175$.}
        \label{fig:ga-IG-vlm90}
    \end{subfigure}
    \vspace*{-5pt}
    \caption{Grounding information gain vs. number of aggregate heads.
    }
    \label{fig:ga-IG}
\end{figure*}
}

\section{LLM Statement}
In this work, large language models (LLMs) are employed in two limited ways: (i) to polish the writing and improve the linguistic clarity of the paper; (ii) to assist in code writing and debugging. 
LLMs are not involved in the design of the core method, the experimental setup, data analysis, or the interpretation of the results.
All texts presented in the paper, as well as the code, are endorsed by the authors, and the authors take full responsibility of the content presented in this paper. 
%%%%%%%%%%%%%%%%%%%%%%%%%%%%%%%%%%%%%%%%%%%%%%%%%%%%%%%%%%%%%%%%%%%%%%%%%%%%%%%
%%%%%%%%%%%%%%%%%%%%%%%%%%%%%%%%%%%%%%%%%%%%%%%%%%%%%%%%%%%%%%%%%%%%%%%%%%%%%%%

\end{document}